\pdfoutput=1
\documentclass[10pt,twocolumn,letterpaper]{article}

\usepackage[pagenumbers]{cvpr}

\usepackage{amsmath, amssymb, amsfonts, amsthm}
\usepackage{mathtools}
\usepackage{bm}

\usepackage{graphicx}
\usepackage{subcaption}
\usepackage{booktabs}
\usepackage{multirow}
\usepackage{array}
\usepackage{makecell}
\usepackage{xcolor}
\usepackage{colortbl}
\usepackage{tabularx}

\usepackage{algorithm}
\usepackage{algorithmic}

\usepackage{url}
\usepackage{verbatim}
\usepackage{enumitem}
\usepackage{xspace}
\usepackage{microtype}

\newcolumntype{C}[1]{>{\centering\arraybackslash}p{#1}}
\newcolumntype{L}[1]{>{\raggedright\arraybackslash}p{#1}}
\newcolumntype{R}[1]{>{\raggedleft\arraybackslash}p{#1}}

\graphicspath{{Figure/}{figures/}{./}}

\usepackage{graphicx}
\usepackage{booktabs}
\usepackage{xcolor}
\usepackage{colortbl}
\usepackage{placeins}
\usepackage{dblfloatfix}

\definecolor{tabNative}{RGB}{220,220,220}

\definecolor{gainpos}{RGB}{0,125,60}
\definecolor{gainneg}{RGB}{190,30,30}

\providecommand{\tPH}{%
    \textcolor{black!35}{--}%
}

\providecommand{\tNA}{%
    \textcolor{black!45}{N/A}%
}

\providecommand{\tnat}[1]{%
    \cellcolor{tabNative}#1%
}

\providecommand{\mstd}[2]{%
    #1\hspace{0.15em}{\tiny (#2)}%
}

\providecommand{\twin}[2]{%
    \hspace{0.35em}{\scriptsize (#1/#2)}%
}

\providecommand{\up}[1]{\textcolor{gainpos}{#1}}
\providecommand{\dn}[1]{\textcolor{gainneg}{#1}}

\providecommand{\sg}[1]{$^{#1}$}

\providecommand{\Dall}{\textsc{Full}}
\providecommand{\Dann}{\textsc{Ann}}

\newcommand{\scout}{Scout\xspace}

\DeclareRobustCommand{\appref}[1]{\cref{#1}}

\definecolor{cvprblue}{rgb}{0.21,0.49,0.74}

\usepackage[
    pagebackref,
    breaklinks,
    colorlinks,
    allcolors=cvprblue
]{hyperref}

\title{One Prior, Every Encoder: Distilling Where to Look into Histopathology MIL}

\author{
Juhyeon Kim \quad Chanjae Song \quad Mun Yong Yi$^{*}$\\[2pt]
Korea Advanced Institute of Science and Technology (KAIST), Daejeon, South Korea\\[2pt]
{\tt\small \{wedsed123, chan4535, munyi\}@kaist.ac.kr}\\[2pt]
{\small $^{*}$Corresponding author}
}

\AtBeginDocument{%
  \crefname{appendix}{Sec.}{Secs.}%
  \Crefname{appendix}{Section}{Sections}%
}

\begin{document}

\maketitle

\begin{abstract}
Multiple instance learning (MIL) aggregators for whole-slide image (WSI) analysis are typically trained from scratch for each task, often with few labels.
Pretraining aggregators helps, but a pretrained aggregator serves only the patch encoder (PE) it was trained with, while the best PE differs across datasets.
We ask whether a single prior can be reused across tasks and PEs.
Using lesion annotations as an oracle, we show that evidence selection, rather than classification, is the main bottleneck in low-data MIL, identifying attention as the key knowledge to transfer.
We identify two limitations of transferring this attention prior through aggregator weights: \emph{prior--adaptation entanglement}, where freezing prevents task-specific adaptation while fine-tuning erodes useful attention, and \emph{feature-space entanglement}, where pretrained weights remain tied to the feature space of a single PE.
We propose \scout, a \emph{PE-agnostic attention prior} that transfers attention as supervision rather than as weight initialization.
A fixed bank of pathology anchors establishes a shared coordinate system across PEs, in which \scout learns consensus attention from multiple PE-specific teachers.
Once frozen, \scout supervises task-specific aggregators through attention knowledge distillation (AKD), preserving the prior while allowing downstream adaptation.
Without downstream task training, \scout ranks lesion patches more accurately (lesion AUROC) than aggregators trained from scratch on each task.
Across eight benchmarks and 15 PEs, five unseen during pretraining, our method improves over scratch training by 3.1 points on seen PEs and 2.9 on unseen PEs, outperforming existing pretrained aggregators and slide foundation models in most comparisons. A single prior thus transfers across tasks and PEs without retraining or additional inference cost. Code is available at \url{https://github.com/juhyeon-ai/OnePrior}.

\end{abstract}

\section{Introduction}
\label{sec:intro}

\begin{figure}[t]
    \centering
    \includegraphics[width=\linewidth]{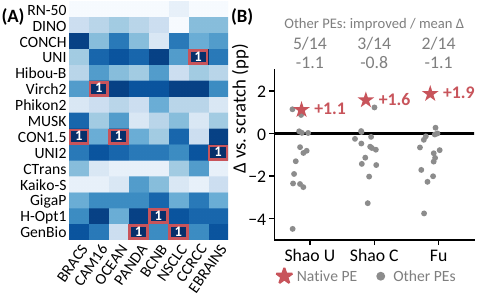}
    \caption{
    \textbf{Feature-space entanglement.}
    \textbf{(A)} PE rank under scratch ABMIL per dataset (darker = better; box = best).
    \textbf{(B)} Gain over scratch from public pretrained ABMIL weights~\cite{shao2025transfer,fu2026pretraining} on their native PE (red) and the 14 other PEs (gray).
    }
    \label{fig:lockin}
\end{figure}

\begin{figure*}[t]
    \centering
    \includegraphics[width=\textwidth]{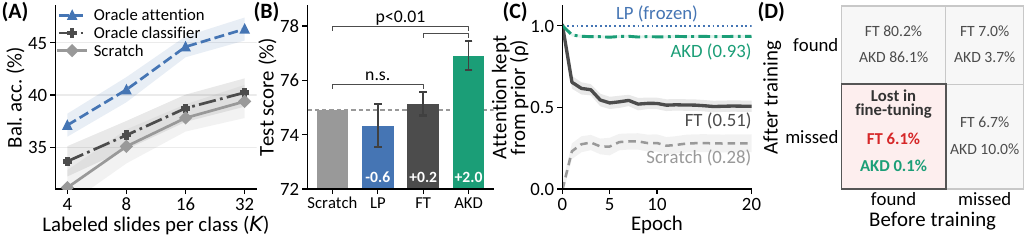}
    \caption{
    \textbf{Prior--adaptation entanglement.}
    \textbf{(A)} BRACS, $K$ slides per class, mean over 15 PEs. \emph{Oracle attention}: attention fixed to lesion masks. \emph{Oracle classifier}: attention frozen after $K$ slides, classifier trained on all remaining slides.
    \textbf{(B--D)} ABMIL pretrained as in~\cite{shao2025transfer} on TCGA and HistAI, native PE, averaged over {\Dann} and 10 PEs; AKD: the frozen pretrained ABMIL supervises the attention of an aggregator trained from scratch.
    \textbf{(B)} Test score; numbers: difference from scratch.
    \textbf{(C)} Rank correlation with the pretrained attention.
    \textbf{(D)} Lesion-bearing slides (\%) by whether the prior located the lesion before training and the model still does after.
    }
    \label{fig:entangle}
\end{figure*}

Whole-slide images (WSIs) are commonly analyzed using a frozen patch encoder (PE) and a multiple instance learning (MIL) aggregator that combines informative patch evidence using slide-level labels~\cite{ilse2018attention,shao2021transmil}.
While pretrained foundation-model PEs are routinely reused across datasets~\cite{chen2024uni,ding2025titan,zimmermann2024virchow2}, MIL aggregators are typically trained from scratch for each task.
This makes it difficult to learn reliable aggregation strategies in low-data settings, such as rare diseases, cancer subtyping, and biomarker prediction.
Recent studies show that MIL aggregators pretrained through pan-cancer classification~\cite{shao2025transfer} or knowledge distillation from slide foundation models~\cite{fu2026pretraining} outperform scratch training, with benefits that persist across organs and tasks~\cite{shao2025transfer}.
We refer to such task-transferable knowledge as a \emph{prior}.
However, each pretrained aggregator is tied to its pretraining PE, requiring a separate checkpoint per PE~\cite{shao2025transfer}, whereas the best PE differs from dataset to dataset (\cref{fig:lockin}A) and is selected per dataset~\cite{kim2026sam}.
Released weights applied to another PE perform comparably to or worse than random initialization (\cref{fig:lockin}B).
This raises our central question: can a single prior be reused across PEs?
Answering it requires knowing what the prior consists of and how it should be carried to a new task and a new PE.

We first ask which knowledge is most valuable for downstream learning.
An attention-based aggregator makes two decisions: which patches to attend to, and how to classify the attended evidence.
We separate the two in an oracle experiment using BRACS lesion annotations~\cite{brancati2022bracs}.
Fixing attention to an ideal distribution derived from lesion masks and training only the classifier improves performance by 6.3 points on average over scratch training.
Conversely, learning attention from limited labeled data, freezing it, and training the classifier on all remaining slides yields a much smaller average gain of 1.3 points (\cref{fig:entangle}A).
These findings suggest that evidence selection, rather than classification, is the primary bottleneck in low-data settings.
We therefore regard attention as the key knowledge to transfer through pretraining and term it the \emph{attention prior}.
Existing methods, however, transfer this prior only indirectly, as the weights of a pretrained aggregator.
Weights are a poor carrier of attention for two reasons: downstream training overwrites them, and they are tied to the feature space of one PE.

We call the first problem \emph{prior--adaptation entanglement}.
The attention prior must be adapted to each downstream task, yet it is stored in the same weights that adaptation updates.
To examine how weight-based transfer handles this, we pretrain ABMIL aggregators through pan-cancer classification over 62 diagnostic classes from TCGA and HistAI~\cite{shao2025transfer}.
Freezing pretrained attention while training only the classifier (linear probing, LP) underperforms scratch training (\cref{fig:entangle}B).
Unlike the oracle attention, which helps even when fixed, pretrained attention is imperfect, and freezing it leaves its errors uncorrected.
Fine-tuning the aggregator (FT) can correct them but yields limited gains (\cref{fig:entangle}B).
During FT, pretrained attention changes rapidly in early epochs, reducing its rank correlation with the original attention (\cref{fig:entangle}C), and some previously localized lesions are no longer detected (\cref{fig:entangle}D).
Freezing the weights thus locks in the prior's errors, while updating them erodes its useful parts.

To resolve this dilemma, we propose to transfer the attention prior as supervision rather than as initialization.
The pretrained aggregator stays frozen and supervises the attention of a downstream aggregator trained from scratch, so the prior never resides in the weights being updated and stays intact while the aggregator adapts to the task.
With this attention knowledge distillation (AKD), the downstream aggregator stays close to the pretrained attention (\cref{fig:entangle}C), retains nearly all lesions the prior localized (\cref{fig:entangle}D), and outperforms both scratch training and FT (\cref{fig:entangle}B).
Supervision alone, without any pretrained weights, thus transfers the attention prior more effectively than weight initialization.

\begin{figure*}[t]
    \centering
    \includegraphics[width=\textwidth]{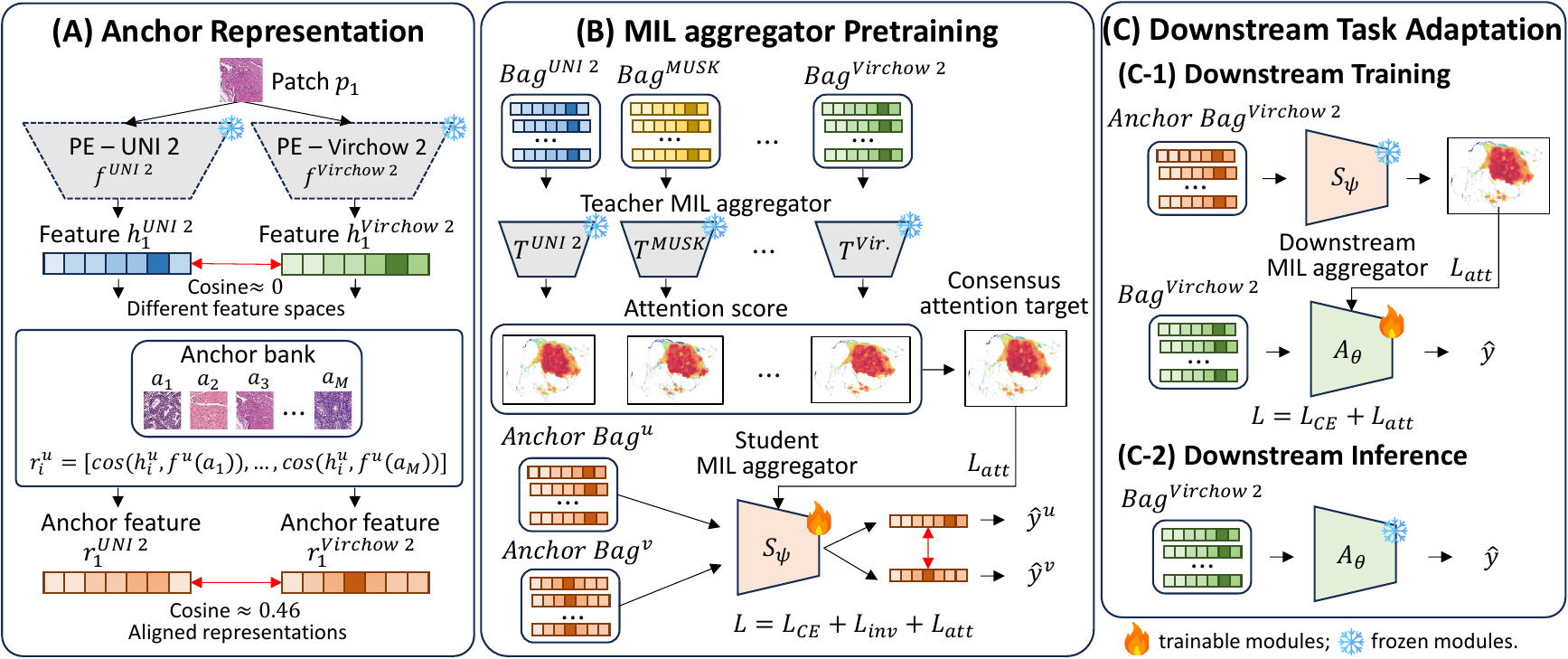}
    \caption{
    \textbf{Overview.}
    \textbf{(A)} A fixed anchor bank re-expresses each patch by its similarities to the anchors, a coordinate system shared by all PEs (\cref{sec:anchor}).
    \textbf{(B)} \scout, an anchor-space student, learns the consensus attention of native-feature teachers (\cref{sec:pretrain}).
    \textbf{(C)} The frozen \scout supervises the attention of a task-specific native-feature aggregator, the only model used at inference (\cref{sec:transfer}).
    }
    \label{fig:method}
\end{figure*}

The frozen prior, however, is still an aggregator that reads the embeddings of one PE.
PEs embed the same patch in different, unaligned feature spaces: their dimensions may differ, and even when they coincide, embeddings of the same patch have near-zero cosine similarity across PEs (\cref{sec:anchor}).
A prior built in one feature space therefore cannot read slides encoded by another PE; applying it would require re-encoding every downstream slide with the pretraining PE, and its attention would still carry that PE's bias.
We call this second problem \emph{feature-space entanglement}.
Attention itself has no such tie: it is a distribution over the patches of a slide and is defined identically under every PE, so a prior expressed as an attention target can cross feature spaces where weights cannot.
Only the input that the prior reads remains PE-specific.
To remove this last dependence, we propose \scout, a \emph{PE-agnostic attention prior}: a single frozen aggregator that supervises attention for any PE, including those unseen during pretraining, without retraining (\cref{fig:method}).
Two design choices enable this.
First, a fixed bank of pathology anchor patches represents each patch through its similarities to the anchors~\cite{moschella2023relative}, establishing a coordinate system shared by all PEs without learning PE-specific mappings.
Supporting a new PE requires only a one-time encoding of the anchor bank.
Second, in this anchor space, \scout is trained as a student aggregator that learns the consensus attention of multiple PE-specific teachers, emphasizing evidence consistently identified across PEs while reducing encoder-specific variation.
Once frozen, \scout supervises task-specific aggregators operating on native PE features through AKD.
The distillation weight decays to zero by mid-training, so the downstream aggregator adapts freely afterwards.

As a result, the prior does not have to be rebuilt when the PE changes.
A weight-based prior for a new PE requires re-encoding all pretraining slides (19{,}766 in our setting) and pretraining a new aggregator, whereas \scout requires encoding the 768 anchor patches once, adds $5$--$10\%$ to downstream training time, and adds no inference cost.

Our contributions are as follows.
\begin{itemize}
\item We identify attention as the key transferable knowledge of pretrained MIL aggregators and show that carrying it in aggregator weights limits its reuse through \emph{prior--adaptation entanglement} and \emph{feature-space entanglement}. We instead transfer it as supervision from a frozen prior, preserving it while the downstream aggregator adapts.

\item We develop \scout, a \emph{PE-agnostic attention prior}. As a student, it learns consensus attention from multiple PE-specific teachers in a shared anchor space, then remains frozen to supervise downstream aggregators through AKD, without retraining for new PEs and without additional inference cost.

\item Across eight benchmarks and 15 PEs, five of them unseen during pretraining, our method improves over scratch training by $+3.1$ points on seen PEs (75 of 80 benchmark--PE pairs) and $+2.9$ on unseen PEs (38 of 40), with larger gains when fewer slides are labeled, and surpasses existing pretrained MIL aggregators and slide foundation models in most comparisons. \scout's zero-shot attention also ranks lesion patches more accurately than task-trained aggregators, with higher lesion AUROC on all four lesion-annotated datasets at matching R-precision on average.
\end{itemize}
\section{Related Work}
\label{sec:related}

\paragraph{Pretraining MIL aggregators.}
Shao \etal~\cite{shao2025transfer} pretrain MIL aggregators~\cite{ilse2018attention,lu2021data,shao2021transmil} through supervised pan-cancer classification, while Fu \etal~\cite{fu2026pretraining} distill slide foundation models. Both transfer pretrained knowledge through aggregator weights, tying the prior to the pretraining PE and exposing it to degradation during fine-tuning. Regularizing fine-tuning toward the pretrained weights~\cite{xuhong2018explicit} limits this degradation but still stores the prior in the weights being updated (\cref{tab:delivery}). In contrast, \scout, our frozen PE-agnostic prior, supervises, rather than initializes, task-specific aggregators on their native PE features.

\paragraph{Slide foundation models.}
Slide foundation models~\cite{wang2024pathology,ding2025titan,vorontsov2026end,zhang2026care} are typically used as frozen slide encoders with linear probing and remain tied to their pretraining PEs. COBRA~\cite{lenz2025cobra} supports unseen PEs by pretraining across multiple PEs, yet relies on PE-specific components and yields a fixed slide embedding. Instead, our single frozen prior supervises trainable aggregators across PEs without PE-specific pretraining of the prior.

\paragraph{Aligning representation spaces.}
Representations learned by different networks can often be related by simple maps: stitching layers connect independently trained models~\cite{lenc2015understanding,bansal2021revisiting}, weight matching and model fusion align their units~\cite{ainsworth2023git,singh2020model}, and latent translation maps one space onto another~\cite{maiorca2023latent}, consistent with similarity analyses indicating that well-trained models learn related structure~\cite{kornblith2019similarity,huh2024platonic}.
These approaches obtain a map for each pair of models, which must be repeated for every new encoder.
Relative representations~\cite{moschella2023relative} instead describe each sample by its similarities to a shared set of anchors, so that independently trained encoders become comparable without a learned map.
We adopt this construction with pathology anchors to give all PEs a common input space for a single MIL prior, and show that the choice of anchors determines how well the prior extends to unseen PEs (\cref{tab:ablation}).

\paragraph{Distillation in MIL.}
Because slide-level labels lack explicit patch-level supervision, several methods use teacher-generated signals to guide MIL learning.
WENO~\cite{qu2022weno} uses bag-level attention scores as pseudo-labels for instance classification, while DAS-MIL~\cite{bontempo2023dasmil} and LRMIL~\cite{shin2026lrmil} distill knowledge across resolutions.
Li \etal~\cite{li2025advancing} preserve attention from earlier models to mitigate forgetting in continual learning.
Attention can also be supervised directly with region annotations~\cite{tourniaire2021mixed,sadafi2023active}, which requires expert labeling for each task.
Unlike these methods, we use AKD to transfer a single pretrained, PE-agnostic prior across tasks and PEs, including unseen ones, without downstream patch-level annotations.

\section{Method}
\label{sec:method}

\paragraph{Design requirements.}
A prior that is reused across tasks and PEs must meet three requirements, and each of them determines one component of our method (\cref{fig:method}).
(i)~\emph{Readable under any PE.}
A single model must accept slides encoded by any PE, including PEs released after pretraining, without learning a PE-specific map.
We therefore re-express every patch in a coordinate frame defined by a fixed bank of pathology anchors (\cref{sec:anchor}).
(ii)~\emph{Accurate despite a shared input.}
A shared coordinate frame is a lossy view of each PE, so a model trained directly on it starts from a degraded input.
We therefore obtain the attention from teachers trained on native features and train an anchor-space student on their consensus (\cref{sec:pretrain}).
(iii)~\emph{Preserved yet adaptable.}
The prior must survive downstream training without preventing the aggregator from adapting to the task.
We therefore keep the student frozen and deliver its attention as a supervision target whose weight decays during training (\cref{sec:transfer}).

\paragraph{Notation.}
A WSI is a bag of $N$ patches $\{p_i\}_{i=1}^{N}$ with slide label $y$.
A frozen PE $f^{v}$, $v\in\mathcal{V}$, maps each patch to $\mathbf{h}^{v}_i=f^{v}(p_i)\in\mathbb{R}^{d_v}$; the superscript $v$ marks quantities of PE $v$.
Let $\mathcal{V}_{\mathrm{seen}}\subset\mathcal{V}$ denote the PEs used in student pretraining, and $\mathcal{T}\subset\mathcal{V}_{\mathrm{seen}}$ the subset used to train teachers.
$\boldsymbol{\alpha}\in\Delta^{N-1}$ denotes the aggregator's attention over patches.

\subsection{A PE-agnostic coordinate frame}
\label{sec:anchor}

\begin{figure}[t]
    \centering
    \includegraphics[width=\linewidth]{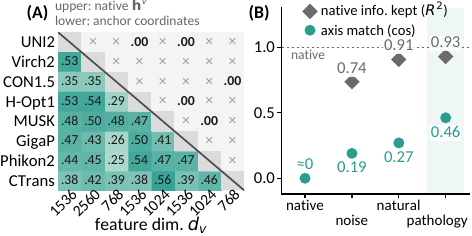}
    \caption{
    \textbf{Anchor coordinates align PE feature spaces.}
    \textbf{(A)} Same-patch cosine across 8 of the 15 PEs (all pairs in \appref{app:cosine}). Upper: native features ($\times$: dimensions differ); lower: anchor coordinates.
    \textbf{(B)} Axis match (same-patch cosine, mean over PE pairs) and native information kept ($R^2$ of a linear map from coordinates to native features), for 768 noise, natural-image, and pathology anchors.
    }
    \label{fig:anchor}
\end{figure}

We fix $M=768$ anchor patches
$\mathcal{A}=\{a_k\}_{k=1}^{M}$, drawn from
TCGA, HistAI, and SPIDER using class-balanced sampling,
as detailed in \appref{app:anchor}.
Following relative representations~\cite{moschella2023relative},
the patch feature $\mathbf{h}^{v}_i$ under PE $v$
is re-expressed by its cosine similarities to anchor
embeddings (\cref{fig:method}A),
\begin{equation}
\mathbf{r}^{v}_i=\big[\,\cos\big(\mathbf{h}^{v}_i,f^{v}(a_1)\big),\;\dots,\;\cos\big(\mathbf{h}^{v}_i,f^{v}(a_M)\big)\,\big]\in\mathbb{R}^{M}.
\label{eq:relrep}
\end{equation}
We call the bag $\{\mathbf{r}^{v}_i\}_{i=1}^{N}$ an \emph{anchor bag}.
A single aggregator cannot read native feature spaces: their dimensions differ, and even when they coincide the coordinates of the same patch are unaligned across PEs (\cref{fig:anchor}A, upper; cosine $0.00$).
A learned map between them~\cite{bansal2021revisiting,ainsworth2023git,maiorca2023latent} must be retrained for every new PE.
In anchor coordinates, the $k$-th axis means ``similarity to anchor $a_k$'' under every PE, giving the representations of the same patch a shared coordinate semantics across PEs (\cref{fig:anchor}A, lower; same-patch cosine $0.46$), and a new PE is supported by encoding $\mathcal{A}$ once.
Domain-relevant anchors are important for effective alignment: natural-image anchors keep nearly as much native information ($0.91$ vs.\ $0.93$) but align the axes far less ($0.27$ vs.\ $0.46$), and noise anchors are worse on both ($0.74$ and $0.19$; \cref{fig:anchor}B).

\subsection{Teacher--student attention prior pretraining}
\label{sec:pretrain}

The \emph{student} $S_\psi$, the aggregator that holds the prior, takes anchor bags $\{\mathbf{r}^{v}_i\}_{i=1}^{N}$ as input so that it applies to any PE.
Anchor coordinates, however, are a lossy view of the native features (\cref{fig:anchor}B), and slide labels alone do not tell an aggregator which patches to attend to.
A student trained from labels on anchor bags would therefore learn its attention from a weak signal on a degraded input.
Native features carry more information, but an aggregator trained on them is bound to one PE.
We therefore separate two roles (\cref{fig:method}B):
\emph{teachers} trained on native features decide
where the evidence is, and the student learns to
reproduce their attention from anchor bags.
Attention is a distribution over the patches of a slide, so it is defined identically in every feature space and passes from native-feature teachers to an anchor-space student without a map between them.
We first train the teachers (\cref{sec:teachers}), then the student on their consensus (\cref{sec:student}).

\subsubsection{Training the teachers}
\label{sec:teachers}

For each teacher PE $t\in\mathcal{T}$, we train a
gated-attention aggregator $T^{t}$ on native features
$\{\mathbf{h}^{t}_i\}_{i=1}^{N}$ using slide-level
diagnostic labels.
Slide-level classification rewards attention only on the patches that best separate the pretraining classes, and a few highly discriminative patches are often enough to do so.
A prior, however, must serve tasks that are unknown at pretraining, such as grading or biomarker prediction, whose evidence may lie in tissue that the pretraining labels do not require.
We therefore regularize teacher attention to cover diverse class-relevant evidence rather than a few highly discriminative patches, using the attention regularizers of ACMIL~\cite{zhang2024attention} and AttriMIL~\cite{cai2025attrimil} (\appref{app:teacher}).
The gap between classification and evidence selection also appears during training: a teacher's attention precision on annotated validation slides peaks near the first epoch and then declines while its cross-entropy keeps decreasing (\cref{fig:supp_ckpt}A).
We therefore stop each teacher near its attention-precision peak rather than at its lowest classification loss (\appref{app:impl}).
We train teachers on eight of the ten seen PEs, selected by pretraining validation accuracy; all ten remain student inputs (\appref{app:teacher_selection}).
Teachers are frozen after training and used only to produce attention targets $\boldsymbol{\alpha}^{t}$.

\subsubsection{Training the student}
\label{sec:student}

\begin{figure}[t]
    \centering
    \includegraphics[width=\linewidth]{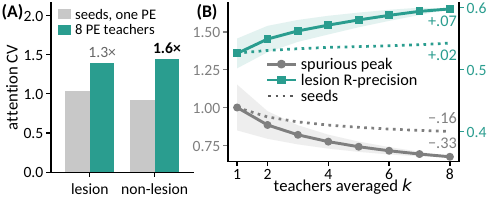}
    \caption{
    \textbf{PE-specific disagreement between teachers is larger outside lesions.}
    Test slides of \Dann.
    \textbf{(A)} Per-patch coefficient of variation (CV) of attention across seeds of one PE and across PE teachers.
    \textbf{(B)} Spurious peak (attention on the most attended non-lesion patches, relative to $k{=}1$) and lesion R-precision when $k$ teachers are averaged; dotted: $k$ seeds of one PE; bands: s.d.\ across~subsets.
    }
    \label{fig:consensus}
\end{figure}

Our goal is to learn a single attention prior that captures
evidence consistently identified across PEs, rather than
attention patterns specific to any one encoder.
A natural target is the attention of a single
PE-specific teacher, but it carries PE-specific bias.
Slide labels indicate what to predict without explicitly
constraining attention on non-lesion patches, allowing
teachers to assign residual attention to different
irrelevant regions.
Indeed, on unseen test slides of \Dann{}, attention
disagreement is greater across PE teachers than across
seeds of the same PE, with a larger relative increase
on non-lesion than lesion patches
($1.58\times$ vs.\ $1.34\times$, $p<0.001$;
\cref{fig:consensus}A).
A single teacher would thus transfer its PE-specific
spurious attention into the shared prior.

To reduce this bias, we form a consensus across
teachers trained in different PE feature spaces.
Averaging all eight teachers reduces non-lesion spurious
peaks by 33\% and increases lesion R-precision from
0.53 to 0.60, exceeding the best single teacher
selected post hoc per cohort (0.55).
In contrast, averaging the same number of seeds
of one PE reduces spurious peaks by only 16\%
and improves R-precision by 0.02
(\cref{fig:consensus}B).
These results suggest that PE diversity provides substantial
benefits beyond ensembling alone.

We therefore define the consensus target as
$\bar{\boldsymbol{\alpha}}=\tfrac{1}{|\mathcal{T}|}\sum_{t\in\mathcal{T}}\boldsymbol{\alpha}^{t}$.
The student is trained on anchor bags from all seen PEs,
with $\boldsymbol{\alpha}^{S}=
S_\psi(\{\mathbf{r}^{v}_i\}_{i=1}^{N})$
denoting its attention under PE $v$.
We match its attention to the consensus using
$\mathcal{L}_{\mathrm{att}}
(\boldsymbol{\alpha},\bar{\boldsymbol{\alpha}})
=\mathrm{KL}(\bar{\boldsymbol{\alpha}}\|\boldsymbol{\alpha})$~\cite{li2025advancing}.
Because anchor coordinates are not perfectly aligned
across PEs (\cref{sec:anchor}), we additionally enforce
consistency between two seen-PE views $v,u$ of the same
slide through an invariance term $\mathcal{L}_{\mathrm{inv}}$
that aligns student attention and slide embeddings
(\appref{app:student}):
\begin{equation}
\mathcal{L}_S =
\mathcal{L}_{\mathrm{CE}} +
\mathcal{L}_{\mathrm{att}}
\big(\boldsymbol{\alpha}^{S},
\bar{\boldsymbol{\alpha}}\big)
+ \gamma\,\mathcal{L}_{\mathrm{inv}}.
\label{eq:student}
\end{equation}
On the patches shared by the two views, $\mathcal{L}_{\mathrm{inv}}$ combines a symmetric InfoNCE loss between the student's patch embeddings, a squared distance between its standardized attention scores, and a cosine distance between its slide embeddings (\cref{eq:supp_inv}).
It thereby encourages the student to produce consistent attention and representations for a slide regardless of the PE that encoded it, a property that is desirable for generalization to unseen PEs.
After pretraining, the student is frozen; we refer to the frozen student $S_\psi$ as \scout.

\subsection{Transferring the prior to downstream tasks}
\label{sec:transfer}

To preserve the pretrained attention prior while allowing
task-specific adaptation, we keep \scout frozen and
use its attention to guide a trainable downstream aggregator
through AKD.
For a downstream task with PE $v\in\mathcal{V}$,
seen or unseen, we train a new aggregator $A_\theta$
from scratch on native features
$\{\mathbf{h}^{v}_i\}_{i=1}^{N}$,
while \scout provides attention targets
from the anchor bag of the same slide (\cref{fig:method}C-1):
\begin{equation}
\mathcal{L}_A =
\mathcal{L}_{\mathrm{CE}}
+ \beta(e)\,
\mathcal{L}_{\mathrm{att}}
\big(\boldsymbol{\alpha}^{A}, \boldsymbol{\alpha}^{S}\big),
\label{eq:transfer}
\end{equation}
where $\boldsymbol{\alpha}^{A}$ is the attention of $A_\theta$,
$\boldsymbol{\alpha}^{S}=
S_\psi(\{\mathbf{r}^{v}_i\}_{i=1}^{N})$
is the attention of \scout, and $e$ is the epoch.

This arrangement divides the two decisions of an attention-based aggregator between the two models.
\scout decides \emph{where to look}: it reads anchor coordinates, which are shared across PEs but lossy, yet retain sufficient information for effective lesion-patch ranking (\cref{sec:rq1}).
$A_\theta$ learns \emph{what to read} there: it operates on native features, which retain the information that anchor coordinates discard (\cref{fig:anchor}B), and its classifier is trained on the downstream labels.
Using \scout itself as the downstream aggregator would give up both advantages: it would classify from lossy anchor coordinates, and updating its weights would erode the attention it was pretrained to provide (\cref{tab:delivery,fig:lost}).

The prior is most valuable early in training, when the classifier
is untrained and slide-level supervision provides
limited guidance on where to attend, consistent with evidence
that regularization shapes deep networks mainly during early
training~\cite{achille2017critical,golatkar2019time}.
As the classifier learns, the task loss can refine attention
beyond the prior, whereas a constant distillation weight
would constrain adaptation toward the task optimum.
We therefore decay the distillation weight linearly to zero
over the first half of training:
\begin{equation}
\beta(e)=\beta_0\max\!\big(0,\;1-2e/E\big),
\label{eq:beta}
\end{equation}
where $E$ is the total number of training epochs.
This allows the aggregator to initially benefit from the
pretrained evidence-selection prior and subsequently adapt
freely to the downstream task.
The decay endpoint is not critical (\cref{tab:delivery}).

At inference, only $A_\theta$ is used
(\cref{fig:method}C-2), introducing no additional test-time cost.
During training, \scout's attention targets are computed once
per slide and cached, so our method takes only $1.05$--$1.10\times$
the training time of scratch ABMIL (\appref{app:compute}).
Thus, our method differs from scratch training only by
an AKD term that vanishes by mid-training,
while a single frozen \scout can supervise aggregators
across downstream tasks and PEs.
\section{Experiments}
\label{sec:experiments}

\subsection{Experimental setup}
\label{sec:setup}

\paragraph{Pretraining.}
The teachers and student are pretrained on 19{,}766 labeled WSIs from TCGA~\cite{weinstein2013cancer} and HistAI~\cite{nechaev2025histai}, spanning 62 diagnostic classes across 23 organs; details are in \appref{app:pretraining}. Pretraining checkpoints are selected by attention precision on TCGA validation slides with public tumor annotations~\cite{loeffler2021tcga}; these annotations enter no training loss, and the annotated slides are disjoint from all downstream benchmarks (\appref{app:impl}). We evaluate 15 PEs listed in \cref{tab:bracs}, comprising 10 \emph{seen} PEs whose features are used during pretraining and 5 \emph{unseen} PEs excluded from all pretraining stages; their details are in \appref{app:encoder}.

\paragraph{Downstream.}
We evaluate twelve tasks on eight benchmarks spanning detection, subtyping, grading, and biomarker prediction over seven organs: BRACS~\cite{brancati2022bracs}, CAMELYON16~\cite{ehteshami2017diagnostic}, UBC-OCEAN~\cite{asadi2024ocean}, PANDA~\cite{bulten2022artificial}, BCNB~\cite{xu2021predicting}, CPTAC NSCLC~\cite{gillette2020proteogenomic,satpathy2021proteogenomic}, CPTAC CCRCC~\cite{clark2019integrated}, and EBRAINS~\cite{roetzer2022digital}; details are in \appref{app:downstream}. We refer to this suite as \Dall. Four of its datasets (CAMELYON16, BRACS, UBC-OCEAN, PANDA) release region annotations, which we never use for training and use only to evaluate attention; we refer to these four as \Dann. \emph{Downstream score} is the task metric averaged over 5 seeds and, unless stated otherwise, over the eight benchmarks. \emph{Lesion AUROC} is the per-slide AUROC of a model's patch attention between lesion and non-lesion patches of \Dann, averaged over test slides. \emph{Lesion R-precision} is the fraction of lesion patches among the $R$ most attended patches, with $R$ the number of lesion patches of the slide.

\paragraph{MIL architecture.}
Unless otherwise specified, the teachers, the student, and the downstream aggregators are gated-attention ABMIL~\cite{ilse2018attention}.
In figures and tables, \emph{Ours} denotes a downstream aggregator trained with AKD from \scout, and \emph{Ours (zero-shot)} denotes \scout itself, evaluated without task training.
The scratch baseline uses the same downstream architecture and training protocol without AKD, isolating the effect of the prior.
The rationale for choosing ABMIL and the implementation details are in \appref{app:impl}.

\subsection{Does the prior localize evidence zero-shot?}
\label{sec:rq1}

\begin{figure}[!t]
\centering
\includegraphics[width=\linewidth]{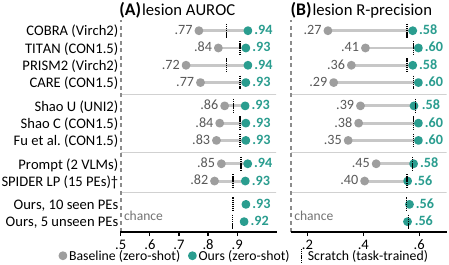}
\caption{
\textbf{Zero-shot localization.}
Lesion localization on \Dann{} of \scout (Ours, teal) and each baseline (gray) on its native PE, without training on \Dann{}.
$^\dagger$Linear probe on SPIDER patch labels, one per PE.
Dotted ticks: task-trained scratch ABMIL, same PE.
}
\label{fig:prior}
\end{figure}

\begin{figure*}[!t]
\centering
\includegraphics[width=\linewidth]{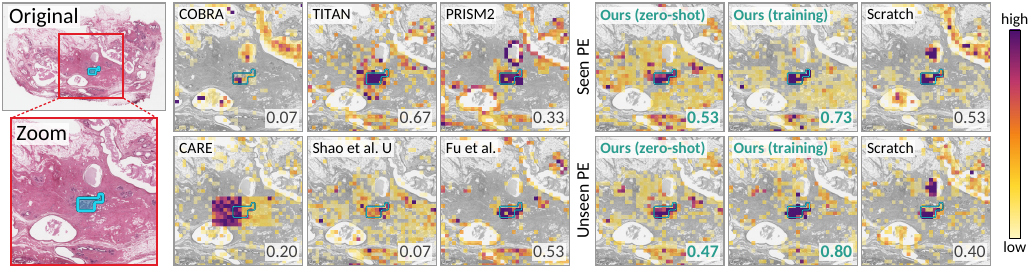}
\caption{
\textbf{Patch attention on a BRACS slide.}
Cyan: annotated region of interest. Left: existing pretrained models on native PEs. Right: \scout (Ours, zero-shot), the aggregator trained with it (Ours, training), and scratch ABMIL under seen (UNI-2) and unseen (GigaPath) PEs.
Numbers: lesion R-precision.
}
\label{fig:qual}
\end{figure*}

\begin{table*}[!t]
\centering
\caption{
\textbf{BRACS 7-class subtyping across 15 PEs.}
Balanced accuracy (\%), s.d.\ in parentheses.
Gray: native PE(s) of each method (ours: the 10 seen PEs).
Green/red: above/below scratch ABMIL.
Bold: best per PE.
}
\label{tab:bracs}

\scriptsize
\setlength{\tabcolsep}{2.4pt}
\renewcommand{\arraystretch}{1.07}

\resizebox{\textwidth}{!}{%
\begin{tabular}{
    l
    *{10}{c}
    @{\hspace{3pt}}
    *{5}{c}
}

\toprule

& \multicolumn{10}{c}{Seen during our pretraining}
& \multicolumn{5}{c}{Unseen} \\

\cmidrule(lr){2-11}
\cmidrule(lr){12-16}

Method
& \thead{RN-50\\\cite{he2016deep}}
& \thead{DINO\\\cite{caron2021emerging}}
& \thead{CONCH\\\cite{lu2024visual}}
& \thead{UNI\\\cite{chen2024uni}}
& \thead{Hibou-B\\\cite{nechaev2024hibou}}
& \thead{Virch2\\\cite{zimmermann2024virchow2}}
& \thead{Phikon2\\\cite{filiot2024phikonv2}}
& \thead{MUSK\\\cite{xiang2025musk}}
& \thead{CON1.5\\\cite{ding2025titan}}
& \thead{UNI2\\\cite{chen2024uni}}
& \thead{CTrans\\\cite{wang2022ctranspath}}
& \thead{Kaiko-S\\\cite{ai2024largescale}}
& \thead{GigaP\\\cite{xu2024gigapath}}
& \thead{H-Opt1\\\cite{scalbert2026hoptimus1}}
& \thead{GenBio\\\cite{kapse2026genbiopathfm}} \\

\midrule

\multicolumn{16}{l}{\textit{Baseline MIL}} \\

DTFD-MIL~\cite{zhang2022dtfd}
& \mstd{35.3}{4.7}
& \mstd{38.8}{6.1}
& \mstd{47.3}{3.9}
& \mstd{37.9}{4.3}
& \mstd{39.5}{4.4}
& \mstd{41.4}{2.5}
& \mstd{39.7}{3.2}
& \mstd{40.9}{5.4}
& \mstd{45.5}{4.1}
& \mstd{44.7}{2.8}
& \mstd{40.6}{4.6}
& \mstd{37.0}{5.4}
& \mstd{38.8}{5.8}
& \mstd{42.6}{4.1}
& \mstd{40.9}{3.4}
\\

TransMIL~\cite{shao2021transmil}
& \mstd{29.6}{3.1}
& \mstd{37.2}{3.5}
& \mstd{41.8}{3.0}
& \mstd{33.3}{3.7}
& \mstd{31.1}{3.4}
& \mstd{35.7}{3.7}
& \mstd{29.5}{3.8}
& \mstd{35.9}{4.0}
& \mstd{43.3}{5.7}
& \mstd{37.0}{5.6}
& \mstd{36.4}{5.5}
& \mstd{35.7}{3.7}
& \mstd{31.0}{4.3}
& \mstd{33.9}{5.0}
& \mstd{39.3}{4.1}
\\

WiKG~\cite{li2024dynamic}
& \mstd{33.8}{2.2}
& \mstd{30.1}{4.9}
& \mstd{46.2}{4.4}
& \mstd{35.7}{3.0}
& \mstd{33.7}{5.0}
& \mstd{39.9}{3.2}
& \mstd{33.1}{3.5}
& \mstd{37.1}{6.3}
& \mstd{43.9}{3.2}
& \mstd{40.4}{2.7}
& \mstd{37.6}{4.0}
& \mstd{28.7}{8.8}
& \mstd{36.0}{2.8}
& \mstd{41.0}{4.1}
& \mstd{44.1}{3.7}
\\

ILRA~\cite{xiang2023ilra}
& \mstd{32.0}{4.2}
& \mstd{34.8}{4.0}
& \mstd{43.1}{3.5}
& \mstd{32.3}{3.4}
& \mstd{38.3}{2.3}
& \mstd{39.7}{3.6}
& \mstd{28.4}{6.4}
& \mstd{41.6}{3.9}
& \mstd{45.1}{6.4}
& \mstd{39.5}{5.3}
& \mstd{38.9}{2.9}
& \mstd{36.9}{5.3}
& \mstd{31.6}{3.3}
& \mstd{36.5}{2.8}
& \mstd{40.2}{5.9}
\\

\midrule

\multicolumn{16}{l}{\textit{Slide Foundation Models}} \\

COBRA~\cite{lenz2025cobra}
& \mstd{31.2}{4.0}
& \mstd{43.9}{1.8}
& \mstd{41.0}{13.1}
& \tnat{\mstd{25.5}{0.6}}
& \mstd{41.2}{0.9}
& \tnat{\mstd{28.6}{7.0}}
& \mstd{39.5}{2.2}
& \mstd{42.3}{6.7}
& \mstd{43.8}{7.0}
& \mstd{40.8}{2.9}
& \tnat{\mstd{33.2}{4.0}}
& \mstd{35.9}{6.0}
& \mstd{38.5}{4.2}
& \mstd{41.5}{6.3}
& \mstd{40.6}{2.8}
\\

TITAN~\cite{ding2025titan}
& \mstd{29.9}{2.7}
& \mstd{41.9}{0.4}
& \mstd{44.8}{4.3}
& \mstd{48.5}{3.1}
& \mstd{44.0}{7.0}
& \mstd{42.8}{7.6}
& \mstd{35.0}{7.1}
& \mstd{42.1}{4.8}
& \tnat{\mstd{43.5}{6.1}}
& \mstd{46.0}{7.9}
& \mstd{41.4}{1.5}
& \mstd{47.0}{0.7}
& \mstd{38.6}{1.6}
& \mstd{40.4}{3.3}
& \mstd{44.5}{9.3}
\\

PRISM2~\cite{vorontsov2026end}
& \mstd{27.2}{3.1}
& \mstd{41.9}{1.3}
& \mstd{49.3}{2.0}
& \mstd{46.6}{5.8}
& \mstd{43.0}{7.0}
& \tnat{\mstd{\textbf{51.8}}{2.5}}
& \mstd{39.0}{4.6}
& \mstd{43.5}{5.8}
& \mstd{43.9}{5.5}
& \mstd{43.8}{3.8}
& \mstd{39.3}{3.1}
& \mstd{41.2}{2.7}
& \mstd{40.1}{1.3}
& \mstd{42.5}{2.7}
& \mstd{41.2}{1.3}
\\

CARE~\cite{zhang2026care}
& \mstd{\textbf{38.8}}{1.5}
& \mstd{\textbf{45.8}}{7.2}
& \mstd{46.6}{3.2}
& \mstd{45.2}{6.6}
& \mstd{34.3}{0.7}
& \mstd{43.5}{6.0}
& \mstd{36.9}{1.3}
& \mstd{38.3}{2.3}
& \tnat{\mstd{40.9}{3.3}}
& \mstd{46.3}{2.6}
& \mstd{41.8}{5.2}
& \mstd{39.5}{2.2}
& \mstd{35.0}{2.8}
& \mstd{41.6}{0.6}
& \mstd{46.4}{7.9}
\\

\midrule

\multicolumn{16}{l}{
    \textit{Pretrained Models (ABMIL)~\cite{ilse2018attention}}
} \\

Scratch
& \mstd{32.4}{3.5}
& \mstd{41.2}{4.8}
& \mstd{50.2}{2.5}
& \mstd{42.8}{4.0}
& \mstd{37.4}{3.4}
& \mstd{41.9}{4.4}
& \mstd{40.4}{3.3}
& \mstd{44.3}{3.8}
& \mstd{49.6}{4.2}
& \mstd{42.7}{3.3}
& \mstd{41.5}{2.5}
& \mstd{38.0}{6.8}
& \mstd{39.7}{2.9}
& \mstd{42.1}{3.0}
& \mstd{43.3}{5.5}
\\

Shao et al.\ U~\cite{shao2025transfer}
& \dn{\mstd{31.7}{2.9}}
& \dn{\mstd{39.1}{2.7}}
& \dn{\mstd{49.7}{2.9}}
& \dn{\mstd{39.1}{2.9}}
& \dn{\mstd{32.6}{4.4}}
& \dn{\mstd{40.5}{3.4}}
& \up{\mstd{41.6}{5.0}}
& \dn{\mstd{41.4}{4.1}}
& \dn{\mstd{48.1}{2.5}}
& \tnat{\up{\mstd{48.4}{4.7}}}
& \dn{\mstd{37.1}{3.2}}
& \up{\mstd{40.0}{3.9}}
& \up{\mstd{42.6}{3.1}}
& \dn{\mstd{40.3}{4.6}}
& \dn{\mstd{42.6}{3.3}}
\\

Shao et al.\ C~\cite{shao2025transfer}
& \dn{\mstd{30.2}{2.6}}
& \up{\mstd{45.0}{4.2}}
& \dn{\mstd{47.9}{2.0}}
& \dn{\mstd{41.9}{3.9}}
& \up{\mstd{42.9}{3.3}}
& \up{\mstd{42.9}{2.5}}
& \up{\mstd{41.9}{3.8}}
& \dn{\mstd{42.2}{2.9}}
& \tnat{\up{\mstd{\textbf{53.2}}{3.8}}}
& \up{\mstd{44.4}{3.5}}
& \dn{\mstd{34.4}{2.8}}
& \up{\mstd{40.1}{5.0}}
& \dn{\mstd{34.1}{4.9}}
& \dn{\mstd{41.8}{3.2}}
& \up{\mstd{43.9}{3.3}}
\\

Fu et al.~\cite{fu2026pretraining}
& \up{\mstd{33.4}{3.5}}
& \dn{\mstd{38.2}{5.7}}
& \dn{\mstd{43.3}{5.8}}
& \dn{\mstd{35.1}{3.0}}
& \dn{\mstd{31.1}{6.5}}
& \dn{\mstd{38.0}{3.7}}
& \dn{\mstd{39.3}{3.8}}
& \dn{\mstd{38.2}{6.3}}
& \tnat{\dn{\mstd{47.5}{3.0}}}
& \dn{\mstd{38.6}{10.1}}
& \dn{\mstd{37.6}{6.2}}
& \dn{\mstd{37.9}{3.2}}
& \dn{\mstd{26.0}{10.2}}
& \dn{\mstd{37.2}{9.5}}
& \dn{\mstd{41.2}{3.3}}
\\

\textbf{Ours}
& \tnat{\up{\mstd{34.1}{2.7}}}
& \tnat{\up{\mstd{\textbf{45.8}}{2.9}}}
& \tnat{\up{\mstd{\textbf{54.1}}{3.0}}}
& \tnat{\up{\mstd{\textbf{49.1}}{1.9}}}
& \tnat{\up{\mstd{\textbf{45.2}}{2.7}}}
& \tnat{\up{\mstd{45.9}{3.7}}}
& \tnat{\up{\mstd{\textbf{47.3}}{1.8}}}
& \tnat{\up{\mstd{\textbf{48.4}}{3.1}}}
& \tnat{\up{\mstd{50.8}{1.9}}}
& \tnat{\up{\mstd{\textbf{49.4}}{2.6}}}
& \up{\mstd{\textbf{43.9}}{2.0}}
& \up{\mstd{\textbf{48.7}}{3.6}}
& \up{\mstd{\textbf{43.9}}{1.1}}
& \up{\mstd{\textbf{44.4}}{1.2}}
& \up{\mstd{\textbf{47.6}}{5.6}}
\\

\bottomrule
\end{tabular}%
}

\end{table*}

AKD can only transfer what its target contains: if \scout's attention does not fall on diagnostic evidence for a task and a PE it was never trained on, distilling it cannot help.
We therefore first examine \scout on its own, before any downstream training.

\paragraph{Localization accuracy.}
We call this evaluation \emph{zero-shot} because \scout never uses the downstream task: none of the slides or labels of \Dann{} were used to train the teachers or the student or to construct the anchor bank, and lesion annotations were never used for training.
Averaged across the four datasets, \scout reaches a lesion AUROC of $0.930$ on seen PEs and $0.924$ on unseen PEs, above ABMIL trained from scratch on each task ($0.886$ and $0.883$; \cref{fig:prior}, bottom rows), at matching lesion R-precision on average ($0.56$ for both models).
On average, \scout thus matches task-trained aggregators on the most attended patches and ranks the remaining patches more accurately.
Its AUROC is higher on all four datasets, by up to $+0.075$ on BRACS, with the smallest margin on CAMELYON16 ($+0.008$); only there, where the task label is lesion presence itself, does the task-trained aggregator reach a higher R-precision (\appref{app:prior_per_dataset}).
The prior is also consistent across PEs: AUROC differs by only $0.006$ between seen and unseen PEs, and \scout's attention rankings on the same slide have a cross-PE rank correlation of $0.92$, including pairs with unseen PEs (\appref{app:consistency}).

\paragraph{Comparison with existing pretrained models.}
Could an existing pretrained model serve as this prior instead?
We compare \scout with nine alternative ways of scoring patches without target-task labels: pretrained MIL aggregators, slide foundation models, text prompts on two vision--language PEs, and a linear probe trained on SPIDER patch labels for each PE, each on its native PE (\cref{fig:prior}; \appref{app:baselines}).
They fall short in two ways.
First, their localization is weaker than that of an aggregator trained from scratch: every baseline falls below task-trained ABMIL in lesion AUROC, and \scout outperforms all nine on matched PEs in both lesion AUROC ($0.93$--$0.94$ vs.\ $0.72$--$0.86$) and lesion R-precision ($0.56$--$0.60$ vs.\ $0.27$--$0.45$).
Second, they do not transfer across PEs: pretrained MIL aggregators and slide foundation models are tied to their native PEs, and the SPIDER linear probe, the only baseline evaluated on all 15 PEs, requires PE-specific training and still trails \scout by $0.11$ AUROC.
Existing pretrained models therefore provide neither the accuracy nor the PE coverage required of a prior.

\paragraph{Qualitative example.}
On a BRACS slide (\cref{fig:qual}), \scout attends to the annotated region under seen and unseen PEs, whereas existing pretrained models spread attention elsewhere, and the aggregator trained with it concentrates further on the annotation; other datasets are in \appref{app:qualitative}.

Without any task training, \scout thus localizes evidence on par with or better than task-trained aggregators, consistently across seen and unseen PEs, which no existing pretrained model achieves.

\subsection{Does one prior improve both seen and unseen PEs?}
\label{sec:main}

\begin{table}[t]
\centering
\caption{
\textbf{Gain over scratch on the eight benchmarks.}
Mean gain over scratch ABMIL on seen and unseen PEs, in points of each metric (0--100 scale); $(k/n)$: PEs improved.
Native: comparisons won against the pretrained methods of \cref{tab:bracs} on their native PEs.
}
\label{tab:summary}

\footnotesize
\setlength{\tabcolsep}{3pt}
\renewcommand{\arraystretch}{1.07}

\begin{tabular*}{\columnwidth}{@{\extracolsep{\fill}} l l c c c}

\toprule

& & \multicolumn{2}{c}{Gain vs.\ scratch} & \\

\cmidrule(lr){3-4}

Benchmark & Metric & Seen & Unseen & Native \\

\midrule

BRACS (2)         & bAcc & +4.2\twin{10}{10} & +3.8\twin{5}{5} & 7/9 \\
CAMELYON16        & Acc  & +2.0\twin{10}{10} & +3.7\twin{4}{5} & 8/9 \\
UBC-OCEAN         & bAcc & +2.1\twin{8}{10}  & +3.1\twin{5}{5} & 6/9 \\
PANDA             & QWK  & +0.8\twin{9}{10}  & +0.9\twin{5}{5} & 8/9 \\
BCNB (3)          & Acc  & +4.0\twin{10}{10} & +3.5\twin{5}{5} & 9/9 \\
CPTAC NSCLC       & Acc  & +5.7\twin{10}{10} & +4.3\twin{5}{5} & 8/9 \\
CPTAC CCRCC       & Acc  & +3.1\twin{9}{10}  & +2.2\twin{4}{5} & 8/9 \\
EBRAINS (2)       & bAcc & +2.7\twin{9}{10}  & +1.5\twin{5}{5} & 5/9 \\

\midrule

All               &       & +3.1\twin{75}{80} & +2.9\twin{38}{40} & 59/72 \\

\bottomrule
\end{tabular*}
\end{table}

For an attention prior to benefit downstream learning, accurate localization must translate into better task performance.
Because the best PE differs across datasets (\cref{fig:lockin}A), a prior is useful only if this gain holds on whichever PE is chosen, including PEs released after pretraining.
We therefore train downstream aggregators with AKD from the same \scout on all 15 PEs and compare them with scratch training and with existing pretrained models.

\paragraph{BRACS across 15 PEs.}
On BRACS seven-class subtyping, existing pretrained models help mainly on their native PEs (\cref{tab:bracs}): applied to other PEs through a trainable linear adapter (\appref{app:baselines}), slide foundation models underperform scratch ABMIL on 7--13 of the 15 PEs, and Shao~U gains $+5.7$ points on its native PE but loses $1.3$ on average on the other 14.
Our method improves over scratch on all 15 PEs, by $+4.7$ and $+4.8$ points for seen and unseen PEs, and ranks first or tied on 12.
Averaged over the 15 PEs, it exceeds the strongest pretrained ABMIL (Shao~C) by $4.8$ points and the strongest slide foundation model (PRISM2) by $4.3$.

\begin{figure}[t]
\centering
\includegraphics[width=\linewidth]{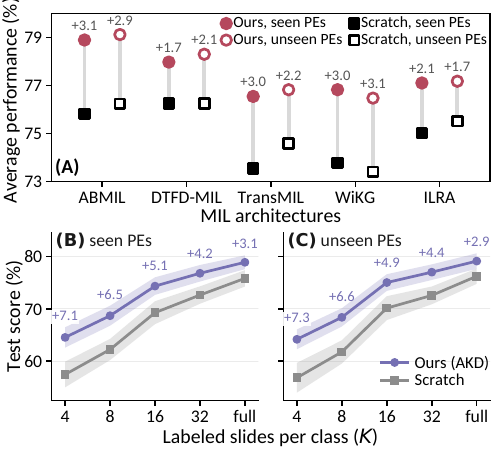}
\caption{
\textbf{Architectures and label budget.}
\textbf{(A)} Downstream score with and without our prior for five MIL architectures.
\textbf{(B, C)} ABMIL trained on $K$ labeled slides per class, seen and unseen PEs; bands: s.d.\ over 5 draws.
Numbers: gain over scratch.
}
\label{fig:generality}
\end{figure}

\paragraph{All benchmarks.}
The gain holds on all eight benchmarks (\cref{tab:summary}; per-PE results in \appref{app:perpe}).
Our method outperforms scratch ABMIL on 75 of 80 seen-PE pairs and 38 of 40 unseen-PE pairs, with nearly equal average gains ($+3.1$ and $+2.9$ points), so the prior loses little on PEs excluded from pretraining.
The gain is smallest on PANDA ($+0.8$/$+0.9$), which has over 6{,}000 training slides, yet 14 of its 15 PEs still improve.
Even on the native PEs of existing pretrained aggregators and slide foundation models, where they are expected to perform best, our method scores higher in 59 of 72 comparisons, with most exceptions on EBRAINS.

\paragraph{Architectures and label budget.}
The prior is not specific to ABMIL: the same AKD improves DTFD-MIL, TransMIL, WiKG, and ILRA by $1.7$--$3.1$ points over scratch on both PE groups (\cref{fig:generality}A).
The gain also grows as labels become scarce, reaching $+7.1$ and $+7.3$ points at $K{=}4$ slides per class for seen and unseen PEs, and remains positive with the full training set (\cref{fig:generality}B,C).

One prior thus improves seen and unseen PEs by nearly the same margin, in 113 of 120 benchmark--PE pairs, across all eight benchmarks and five architectures, and helps most when labels are scarce.

\subsection{How should the prior be delivered?}
\label{sec:preserve}

\begin{table}[t]
\centering
\caption{
\textbf{How the prior is delivered.}
Downstream score of ours on \Dall{} and its change when \scout is delivered differently.
$^{*}$: $p{<}0.05$ (paired Wilcoxon over (dataset, PE)).
}
\label{tab:delivery}
\footnotesize
\setlength{\tabcolsep}{4pt}
\renewcommand{\arraystretch}{1.08}
\begin{tabular*}{\columnwidth}{@{\extracolsep{\fill}} l c c}
\toprule
& \multicolumn{2}{c}{Score / $\Delta$ vs.\ Ours (pts)} \\
\cmidrule(lr){2-3}
Delivery of the prior & Seen PEs & Unseen PEs \\
\midrule
\textbf{Ours}: frozen target, AKD & 78.9 & 79.1 \\
\quad no prior (scratch ABMIL)                  & $-$3.1\sg{*} & $-$2.9\sg{*} \\
\midrule
\multicolumn{3}{@{}l}{\textit{As weights}} \\
\quad initialize from \scout, fine-tune        & $-$2.3\sg{*} & $-$4.5\sg{*} \\
\quad initialize from \scout, linear probe     & $-$8.3\sg{*} & $-$9.9\sg{*} \\
\quad weight regularization (L2-SP)             & $-$3.7\sg{*} & $-$7.1\sg{*} \\
\midrule
\multicolumn{3}{@{}l}{\textit{As a target other than attention}} \\
\quad slide-embedding distillation              & $-$2.8\sg{*} & $-$4.2\sg{*} \\
\midrule
\multicolumn{3}{@{}l}{\textit{Attention distilled from something else}} \\
\quad uniform attention                         & $-$3.6\sg{*} & $-$5.4\sg{*} \\
\quad shuffled \scout attention                & $-$3.6\sg{*} & $-$5.3\sg{*} \\
\quad attention of a scratch ABMIL (same PE)    & $-$3.0\sg{*} & $-$3.5\sg{*} \\
\midrule
\multicolumn{3}{@{}l}{\textit{Schedule $\beta(e)$ (Ours: linear decay to 0 at $E/2$)}} \\
\quad constant                                  & $-$1.8\sg{*} & $-$1.7\sg{*} \\
\quad first epoch only                          & $-$0.2\sg{} & $-$0.1\sg{} \\
\quad decay over all epochs                     & +0.2\sg{} & $-$0.9\sg{} \\
\bottomrule
\end{tabular*}
\end{table}

The gains above could come from \scout's pretrained knowledge regardless of how it is delivered.
Our design, however, rests on a specific claim: the prior should reach the downstream aggregator as a frozen attention target rather than as weights (\cref{sec:intro}).
To test this claim, we hold \scout fixed and vary only how it reaches the downstream aggregator (\cref{tab:delivery}).

\paragraph{Delivery mechanisms.}
First, the prior must be delivered as a target, not as weights: fine-tuning \scout, linear probing it, or regularizing toward its weights falls 2--10 points below ours, and on unseen PEs all three fall below training without a prior.
Second, the target must be attention: distilling \scout's slide embedding is $2.8$--$4.2$ points below ours.
Third, the attention must be that of the prior: a uniform target, a shuffled one, or the attention of a scratch ABMIL is no better than no prior.
The schedule matters little (\cref{eq:beta}): only a constant $\beta$ significantly lowers the score.

\begin{figure}[t]
\centering
\includegraphics[width=\linewidth]{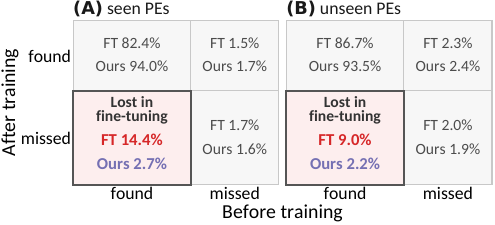}
\caption{
\textbf{What fine-tuning loses and AKD keeps.}
Lesion-bearing slides of \Dann{} (\%) by whether \scout located the lesion before downstream training and the downstream aggregator still does after.
FT: \scout fine-tuned. Ours: AKD from \scout.
}
\label{fig:lost}
\end{figure}

\paragraph{What fine-tuning loses.}
The gap to fine-tuning is accompanied by a substantial difference in prior preservation (\cref{fig:lost}).
Fine-tuning \scout loses lesions it had localized on $14.4\%$ of lesion-bearing slides for seen PEs and $9.0\%$ for unseen PEs, whereas an aggregator trained from scratch with AKD from \scout loses them on only $2.7\%$ and $2.2\%$; both recover a similar share of slides whose lesions \scout had missed ($1.5$--$2.4\%$).
The supervised aggregator also refines the prior: it keeps \scout's lesion AUROC ($0.931$/$0.934$ vs.\ $0.930$/$0.924$ for seen/unseen PEs) and raises lesion R-precision ($0.603$/$0.617$ vs.\ $0.565$/$0.558$; \cref{tab:ablation}).

The prior is therefore best delivered as a frozen attention target: delivering it as weights, as a slide embedding, or with attention from another source loses most or all of the gain, whereas a frozen target preserves what fine-tuning loses while leaving the aggregator free to refine it.

\subsection{What makes the prior work?}
\label{sec:ablation}

\begin{table}[t]
\centering
\caption{
\textbf{Ablations of the prior.}
Zero-shot lesion AUROC and R-precision on \Dann{}; each ablation retrains the student with one component changed.
$^{*}$: $p{<}0.05$ vs.\ Ours (paired Wilcoxon over (dataset, PE), Holm-corrected).
\tNA: not applicable.
}
\label{tab:ablation}
\scriptsize
\setlength{\tabcolsep}{2pt}
\renewcommand{\arraystretch}{1.08}
\begin{tabular*}{\columnwidth}{@{\extracolsep{\fill}} p{0.40\columnwidth} cc cc @{}}
\toprule
& \multicolumn{2}{c}{Lesion AUROC} & \multicolumn{2}{c}{Lesion R-precision} \\
\cmidrule(lr){2-3}\cmidrule(lr){4-5}
& Seen & Unseen & Seen & Unseen \\
\midrule
\textbf{Ours} (zero shot) & 0.930 & 0.924 & 0.565 & 0.558 \\
\textbf{Ours} (after training) & 0.931 & 0.934 & 0.603 & 0.617 \\
Scratch ABMIL & 0.886 & 0.883 & 0.558 & 0.561 \\
\multicolumn{5}{@{}l}{\textit{Rows below: difference from Ours (zero shot)}} \\
\midrule
\multicolumn{5}{@{}l}{\textbf{Q1. What is the prior learned from?} (\cref{sec:setup})} \\
\quad TCGA only  & $-$0.022\sg{*} & $-$0.024\sg{*} & $-$0.065\sg{*} & $-$0.064\sg{*} \\
\quad HistAI only & $-$0.040\sg{*} & $-$0.044\sg{*} & $-$0.081\sg{*} & $-$0.084\sg{*} \\
\midrule
\multicolumn{5}{@{}l}{\textbf{Q2. Does the consensus target matter?} (\cref{sec:teachers,sec:student})} \\
\multicolumn{5}{@{}l}{\textit{Number of teachers (Ours: 8 PEs)}} \\
\quad $|\mathcal{T}|{=}1$ & $-$0.038\sg{*} & $-$0.036\sg{*} & $-$0.043\sg{*} & $-$0.030\sg{*} \\
\quad $|\mathcal{T}|{=}2$ & $-$0.010\sg{*} & $-$0.010\sg{*} & $-$0.028\sg{*} & $-$0.022\sg{*} \\
\quad $|\mathcal{T}|{=}4$ & $-$0.004\sg{*} & $-$0.006\sg{*} & $-$0.019\sg{*} & $-$0.017\sg{*} \\
\quad $|\mathcal{T}|{=}8$, seeds of one PE & $-$0.010\sg{*} & $-$0.008\sg{} & $-$0.022\sg{*} & $-$0.015\sg{*} \\
\multicolumn{5}{@{}l}{\textit{Teacher objective}} \\
\quad w/o $\mathcal{L}_{\mathrm{ACMIL}}$ & $-$0.004\sg{*} & $-$0.003\sg{*} & $-$0.024\sg{*} & $-$0.032\sg{*} \\
\quad w/o $\mathcal{L}_{\mathrm{AttriMIL}}$ & $-$0.020\sg{*} & $-$0.019\sg{*} & $-$0.011\sg{} & $-$0.015\sg{} \\
\quad $\mathcal{L}_{\mathrm{CE}}$ only & $-$0.020\sg{*} & $-$0.020\sg{} & $-$0.031\sg{*} & $-$0.035\sg{*} \\
\midrule
\multicolumn{5}{@{}l}{\textbf{Q3. Does the student need teachers?} (\cref{eq:student})} \\
\quad w/o $\mathcal{L}_{\mathrm{att}}$ (no teacher) & $-$0.054\sg{*} & $-$0.048\sg{*} & $-$0.096\sg{*} & $-$0.096\sg{*} \\
\quad w/o $\mathcal{L}_{\mathrm{CE}}$ & +0.000\sg{} & $-$0.001\sg{} & +0.003\sg{} & $-$0.009\sg{} \\
\quad w/o $\mathcal{L}_{\mathrm{inv}}$ & $-$0.005\sg{*} & $-$0.004\sg{*} & $-$0.020\sg{*} & $-$0.020\sg{*} \\
\quad teacher losses on the student & $-$0.030\sg{*} & $-$0.034\sg{*} & $-$0.066\sg{*} & $-$0.080\sg{*} \\
\midrule
\multicolumn{5}{@{}l}{\textbf{Q4. Are pathology anchors necessary?} (\cref{eq:relrep})} \\
\quad noise anchors & $-$0.104\sg{*} & $-$0.420\sg{*} & $-$0.189\sg{*} & $-$0.389\sg{*} \\
\quad pixel-shuffled anchors & $-$0.047\sg{*} & $-$0.116\sg{*} & $-$0.121\sg{*} & $-$0.223\sg{*} \\
\quad natural-image anchors & $-$0.010\sg{*} & $-$0.075\sg{*} & $-$0.049\sg{*} & $-$0.183\sg{*} \\
\quad learned adapter per PE & $-$0.002\sg{} & \tNA & $-$0.012\sg{} & \tNA \\
\bottomrule
\end{tabular*}
\end{table}

\scout combines four design choices: its pretraining data, the consensus of multiple PE-specific teachers, the supervision of an anchor-space student by native-feature teachers, and pathology anchors.
We ask which of them the prior depends on.
Each variant changes one component of \cref{sec:method}, retrains the student, and is evaluated by zero-shot localization on \Dann{} (\cref{tab:ablation}; variants are defined in \appref{app:ablation}).

\paragraph{Pretraining data and consensus (Q1, Q2).}
Both data sources contribute: using only TCGA or only HistAI lowers lesion AUROC by $0.022$ and $0.040$.
Consensus improves with more teachers through PE diversity, not ensembling: eight seeds of one PE do no better than two distinct PEs ($-0.010$ for both).
Both teacher regularizers matter: ACMIL mainly for lesion R-precision, AttriMIL for lesion AUROC.

\paragraph{Teacher supervision (Q3).}
Removing $\mathcal{L}_{\mathrm{att}}$ causes the largest loss among the student's objective terms ($-0.054$ AUROC, $-0.096$ R-precision), and training the student directly with the teacher losses still leaves a $0.030$ AUROC gap, so an anchor-space student needs native-feature teachers.
$\mathcal{L}_{\mathrm{inv}}$ adds a small gain, and $\mathcal{L}_{\mathrm{CE}}$ has no measurable effect on localization.

\paragraph{Anchor bank (Q4).}
Replacing pathology anchors with natural-image, pixel-shuffled, or noise anchors degrades localization far more on unseen PEs ($-0.075$, $-0.116$, $-0.420$ AUROC) than on seen PEs ($-0.010$, $-0.047$, $-0.104$), suggesting that the student can compensate for poor coordinates on the PEs it is trained with but not on new ones.
A learned linear adapter per PE is no better on seen PEs and cannot serve unseen PEs.

The prior thus derives its localization accuracy from native-feature teacher supervision and cross-PE consensus, while pathology anchors are particularly important for generalization to unseen PEs. Both pretraining data sources further contribute to its accuracy.

\section{Conclusion}
\label{sec:conclusion}

We asked whether a single prior can be reused across tasks and PEs, and \scout shows that it can.
\scout is a PE-agnostic attention prior that learns the consensus attention of multiple PE-specific teachers in a shared pathology anchor space.
Through attention knowledge distillation (AKD), it is transferred as supervision rather than as weight initialization, so a single frozen \scout guides every downstream aggregator while each remains free to adapt to its task.

Without any downstream training, \scout already ranks lesion patches more accurately than aggregators trained on each task.
Aggregators trained with AKD outperform scratch training in 113 of 120 benchmark--PE pairs, with nearly identical gains on seen ($+3.1$ points) and unseen ($+2.9$ points) PEs, and surpass existing pretrained aggregators and slide foundation models in most comparisons, even on their native PEs.
The gains grow to about $7$ points with only four labeled slides per class and carry over to four other MIL architectures.
Beyond accuracy, AKD raises lesion R-precision while retaining \scout's lesion ranking, yielding more interpretable attention maps that align better with annotated lesions, without using any lesion annotation of the downstream task.

These benefits come at minimal cost.
Supporting a new PE takes a one-time encoding of 768 anchor patches, instead of re-encoding the entire pretraining set and pretraining another aggregator.
AKD adds only $5$--$10\%$ to downstream training time and no additional inference cost.

Pretrained attention can thus serve as reusable knowledge that is independent of the choice of PE: one prior accompanies whichever PE suits each dataset best, providing a practical basis for generalizable, data-efficient, computationally efficient, and interpretable WSI analysis.

\paragraph{Limitations.}
Our approach requires patch-level attention and therefore does not directly apply to mean or max pooling.
Distillation may transfer imperfect attention patterns, although downstream adaptation can mitigate their effects.
Pretraining requires multiple PE-specific teachers, incurring a one-time computational cost.
Finally, \scout is pretrained on public data spanning 62 diagnostic classes; broader data coverage, finer-grained labels, or expert annotations may further strengthen the prior.

\section*{LLM Usage Statement}

We used LLM-based tools (ChatGPT and Claude) to correct grammar and refine the wording of author-written text, and to suggest relevant prior work during the early stages of the research. The research ideas, experimental design, manuscript structure, and initial drafts were developed and written by the authors. We also used Claude Code to implement experimental, analysis, and visualization code based on author-provided prompts specifying the intended methods and implementation requirements. All LLM-assisted code and manuscript content were reviewed and verified by the authors, who take full responsibility for the final work.

\section*{Acknowledgments}

This work was supported by the National Research Foundation of Korea (NRF) grant funded by the Korean government (MSIT) under grant number RS-2022-NR068758. The authors also gratefully acknowledge the generous support provided by the Seegene Medical Foundation, Republic of Korea.

{
    \small
    \bibliographystyle{ieeenat_fullname}
    \bibliography{main}

@String(CVPR = {IEEE Conf. Comput. Vis. Pattern Recog.})

@String(ICCV = {Int. Conf. Comput. Vis.})

@String(ECCV = {Eur. Conf. Comput. Vis.})

@String(NIPS = {Adv. Neural Inform. Process. Syst.})

@String(ICLR = {Int. Conf. Learn. Represent.})

@String(MICCAI = {Med. Image Comput. Comput. Assist. Interv.})

@String(MEDIA = {Med. Image Anal.})

@String(NATMED = {Nat. Med.})

@String(NATBME = {Nat. Biomed. Eng.})

@String(NATGEN = {Nat. Genet.})

@String(ICML = {Int. Conf. Mach. Learn.})

@inproceedings{zhang2022dtfd,
  title={{DTFD-MIL}: Double-tier feature distillation multiple instance learning for histopathology whole slide image classification},
  author={Zhang, Hongrun and Meng, Yanda and Zhao, Yitian and Qiao, Yihong and Yang, Xiaoyun and Coupland, Sarah E and Zheng, Yalin},
  booktitle=CVPR,
  pages={18780--18790},
  year={2022}
}

@inproceedings{xiang2023ilra,
  title={Exploring low-rank property in multiple instance learning for whole slide image classification},
  author={Xiang, Jinxi and Zhang, Jun},
  booktitle=ICLR,
  year={2023}
}

@inproceedings{ilse2018attention,
  title={Attention-based deep multiple instance learning},
  author={Ilse, Maximilian and Tomczak, Jakub and Welling, Max},
  booktitle=ICML,
  pages={2127--2136},
  year={2018}
}

@article{shao2021transmil,
  title={{TransMIL}: Transformer based correlated multiple instance learning for whole slide image classification},
  author={Shao, Zhuchen and Bian, Hao and Chen, Yang and Wang, Yifeng and Zhang, Jian and Ji, Xiangyang and others},
  journal=NIPS,
  volume={34},
  pages={2136--2147},
  year={2021}
}

@article{lu2021data,
  title={Data-efficient and weakly supervised computational pathology on whole-slide images},
  author={Lu, Ming Y and Williamson, Drew FK and Chen, Tiffany Y and Chen, Richard J and Barbieri, Matteo and Mahmood, Faisal},
  journal=NATBME,
  volume={5},
  pages={555--570},
  year={2021}
}

@inproceedings{kang2023benchmarking,
  title={Benchmarking self-supervised learning on diverse pathology datasets},
  author={Kang, Mingu and Song, Heon and Park, Seonwook and Yoo, Donggeun and Pereira, S{\'e}rgio},
  booktitle=CVPR,
  pages={3344--3354},
  year={2023}
}

@article{lu2024visual,
  title={A visual-language foundation model for computational pathology},
  author={Lu, Ming Y and Chen, Bowen and Williamson, Drew FK and Chen, Richard J and Liang, Ivy and Ding, Tong and Jaume, Guillaume and Odintsov, Igor and Le, Long Phi and Gerber, Georg and others},
  journal=NATMED,
  volume={30},
  pages={863--874},
  year={2024}
}

@article{ehteshami2017diagnostic,
  title={Diagnostic assessment of deep learning algorithms for detection of lymph node metastases in women with breast cancer},
  author={Ehteshami Bejnordi, Babak and Veta, Mitko and Johannes van Diest, Paul and Van Ginneken, Bram and Karssemeijer, Nico and Litjens, Geert and Van Der Laak, Jeroen AWM and Camelyon16 Consortium and Hermsen, Meyke and Manson, Quirine F and others},
  journal={JAMA},
  volume={318},
  pages={2199--2210},
  year={2017}
}

@article{brancati2022bracs,
  title={{BRACS}: A dataset for breast carcinoma subtyping in {H\&E} histology images},
  author={Brancati, Nadia and Anniciello, Anna Maria and Pati, Pushpak and Riccio, Daniel and Scognamiglio, Giosu{\`e} and Jaume, Guillaume and De Pietro, Giuseppe and Di Bonito, Maurizio and Foncubierta, Antonio and Botti, Gerardo and others},
  journal={Database},
  volume={2022},
  pages={baac093},
  year={2022}
}

@article{weinstein2013cancer,
  title={The {Cancer Genome Atlas} {Pan-Cancer} analysis project},
  author={Weinstein, John N and Collisson, Eric A and Mills, Gordon B and Shaw, Kenna R and Ozenberger, Brad A and Ellrott, Kyle and Shmulevich, Ilya and Sander, Chris and Stuart, Joshua M},
  journal=NATGEN,
  volume={45},
  pages={1113--1120},
  year={2013}
}

@article{bulten2022artificial,
  title={Artificial intelligence for diagnosis and {Gleason} grading of prostate cancer: the {PANDA} challenge},
  author={Bulten, Wouter and Kartasalo, Kimmo and Chen, Po-Hsuan Cameron and Str{\"o}m, Peter and Pinckaers, Hans and Nagpal, Kunal and Cai, Yuannan and Steiner, David F and Van Boven, Hester and Vink, Robert and others},
  journal=NATMED,
  volume={28},
  pages={154--163},
  year={2022}
}

@inproceedings{shao2025transfer,
  title={Do Multiple Instance Learning Models Transfer?},
  author={Shao, Daniel and Chen, Richard J and Song, Andrew H and Runevic, Joel and Lu, Ming Y and Ding, Tong and Mahmood, Faisal},
  booktitle=ICML,
  pages={54219--54238},
  year={2025}
}

@inproceedings{kim2026sam,
  title={{SAM-Cluster} and {SAM-LP}: Structure-Aware Evaluation Metrics for Selecting {WSI} Feature Extractors Without {MIL} Training},
  author={Kim, Juhyeon and Ssemakula, Paul and Song, Chanjae and Yi, Mun Yong},
  booktitle=MICCAI,
  pages={473--483},
  year={2026}
}

@article{wang2024pathology,
  title={A pathology foundation model for cancer diagnosis and prognosis prediction},
  author={Wang, Xiyue and Zhao, Junhan and Marostica, Eliana and Yuan, Wei and Jin, Jietian and Zhang, Jiayu and Li, Ruijiang and Tang, Hongping and Wang, Kanran and Li, Yu and others},
  journal={Nature},
  volume={634},
  pages={970--978},
  year={2024}
}

@article{xu2024gigapath,
  title={A whole-slide foundation model for digital pathology from real-world data},
  author={Xu, Hanwen and Usuyama, Naoto and Bagga, Jaspreet and Zhang, Sheng and Rao, Rajesh and Naumann, Tristan and Wong, Cliff and Gero, Zelalem and Gonz{\'a}lez, Javier and Gu, Yu and others},
  journal={Nature},
  volume={630},
  pages={181--188},
  year={2024}
}

@article{vorontsov2026end,
  title={End-to-end multimodal pathology foundation model with clinical dialogue},
  author={Vorontsov, Eugene and Shaikovski, George and Casson, Adam and Viret, Julian and Zimmermann, Eric and Tenenholtz, Neil and Wang, Yi Kan and Bernhard, Jan H and Godrich, Ran A and Retamero, Juan A and others},
  journal=NATMED,
  pages={1--11},
  year={2026}
}

@inproceedings{zhang2026care,
  title={{CARE}: A Molecular-Guided Foundation Model with Adaptive Region Modeling for Whole Slide Image Analysis},
  author={Zhang, Di and Gong, Zhangpeng and Pang, Xiaobo and Liu, Jiashuai and Lu, Junbo and Cui, Hao and Ge, Jiusong and Zeng, Zhi and Yi, Kai and Li, Yinghua and Liu, Si and Yu, Tingsong and Wang, Haoran and Crispin-Ortuzar, Mireia and Yu, Weimiao and Li, Chen and Gao, Zeyu},
  booktitle=CVPR,
  pages={21078--21088},
  year={2026}
}

@inproceedings{lenz2025cobra,
  title={Unsupervised foundation model-agnostic slide-level representation learning},
  author={Lenz, Tim and Neidlinger, Peter and Ligero, Marta and W{\"o}lflein, Georg and Van Treeck, Marko and Kather, Jakob N},
  booktitle=CVPR,
  pages={30807--30817},
  year={2025}
}

@article{cai2025attrimil,
  title={{AttriMIL}: Revisiting attention-based multiple instance learning for whole-slide pathological image classification from a perspective of instance attributes},
  author={Cai, Linghan and Huang, Shenjin and Zhang, Ye and Lu, Jinpeng and Zhang, Yongbing},
  journal=MEDIA,
  volume={103},
  pages={103631},
  year={2025}
}

@inproceedings{zhang2024attention,
  title={Attention-challenging multiple instance learning for whole slide image classification},
  author={Zhang, Yunlong and Li, Honglin and Sun, Yunxuan and Zheng, Sunyi and Zhu, Chenglu and Yang, Lin},
  booktitle=ECCV,
  pages={125--143},
  year={2024}
}

@inproceedings{moschella2023relative,
  title={Relative Representations Enable Zero-shot Latent Space Communication},
  author={Moschella, Luca and Maiorca, Valentino and Fumero, Marco and Norelli, Antonio and Locatello, Francesco and Rodol{\`a}, Emanuele},
  booktitle=ICLR,
  year={2023}
}

@article{maiorca2023latent,
  title={Latent space translation via semantic alignment},
  author={Maiorca, Valentino and Moschella, Luca and Norelli, Antonio and Fumero, Marco and Locatello, Francesco and Rodol{\`a}, Emanuele},
  journal=NIPS,
  volume={36},
  pages={55394--55414},
  year={2023}
}

@inproceedings{lenc2015understanding,
  title={Understanding image representations by measuring their equivariance and equivalence},
  author={Lenc, Karel and Vedaldi, Andrea},
  booktitle=CVPR,
  pages={991--999},
  year={2015}
}

@article{bansal2021revisiting,
  title={Revisiting model stitching to compare neural representations},
  author={Bansal, Yamini and Nakkiran, Preetum and Barak, Boaz},
  journal=NIPS,
  volume={34},
  pages={225--236},
  year={2021}
}

@article{singh2020model,
  title={Model fusion via optimal transport},
  author={Singh, Sidak Pal and Jaggi, Martin},
  journal=NIPS,
  volume={33},
  pages={22045--22055},
  year={2020}
}

@inproceedings{ainsworth2023git,
  title={{Git Re-Basin}: Merging Models modulo Permutation Symmetries},
  author={Ainsworth, Samuel K and Hayase, Jonathan and Srinivasa, Siddhartha},
  booktitle=ICLR,
  year={2023}
}

@inproceedings{huh2024platonic,
  title={The {Platonic} Representation Hypothesis},
  author={Huh, Minyoung and Cheung, Brian and Wang, Tongzhou and Isola, Phillip},
  booktitle=ICML,
  year={2024}
}

@article{fu2026pretraining,
  title={Pretraining Multiple Instance Learning Networks with Multi-Teacher Distillation from Pathology Slide Foundation Models},
  author={Fu, Mingxi and Li, Jiawen and Yan, Renao and Hu, Jiali and Sun, Qiehe and Guan, Tian and He, Yonghong},
  journal={arXiv preprint arXiv:2607.14703},
  year={2026}
}

@inproceedings{li2024dynamic,
  title={Dynamic graph representation with knowledge-aware attention for histopathology whole slide image analysis},
  author={Li, Jiawen and Chen, Yuxuan and Chu, Hongbo and Sun, Qiehe and Guan, Tian and Han, Anjia and He, Yonghong},
  booktitle=CVPR,
  pages={11323--11332},
  year={2024}
}

@inproceedings{he2016deep,
  title={Deep residual learning for image recognition},
  author={He, Kaiming and Zhang, Xiangyu and Ren, Shaoqing and Sun, Jian},
  booktitle=CVPR,
  pages={770--778},
  year={2016}
}

@inproceedings{caron2021emerging,
  title={Emerging properties in self-supervised vision transformers},
  author={Caron, Mathilde and Touvron, Hugo and Misra, Ishan and J{\'e}gou, Herv{\'e} and Mairal, Julien and Bojanowski, Piotr and Joulin, Armand},
  booktitle=ICCV,
  pages={9630--9640},
  year={2021}
}

@article{wang2022ctranspath,
  title={Transformer-based unsupervised contrastive learning for histopathological image classification},
  author={Wang, Xiyue and Yang, Sen and Zhang, Jun and Wang, Minghui and Zhang, Jing and Yang, Wei and Huang, Junzhou and Han, Xiao},
  journal=MEDIA,
  volume={81},
  pages={102559},
  year={2022}
}

@article{chen2024uni,
  title={Towards a general-purpose foundation model for computational pathology},
  author={Chen, Richard J and Ding, Tong and Lu, Ming Y and Williamson, Drew FK and Jaume, Guillaume and Song, Andrew H and Chen, Bowen and Zhang, Andrew and Shao, Daniel and Shaban, Muhammad and others},
  journal=NATMED,
  volume={30},
  pages={850--862},
  year={2024}
}

@article{ai2024largescale,
  title={Towards large-scale training of pathology foundation models},
  author={Aben, Nanne and de Jong, Edwin D and Gatopoulos, Ioannis and K{\"a}nzig, Nicolas and Karasikov, Mikhail and Lagr{\'e}, Axel and Moser, Roman and van Doorn, Joost and Tang, Fei and others},
  journal={arXiv preprint arXiv:2404.15217},
  year={2024}
}

@article{nechaev2024hibou,
  title={Hibou: A family of foundational vision transformers for pathology},
  author={Nechaev, Dmitry and Pchelnikov, Alexey and Ivanova, Ekaterina},
  journal={arXiv preprint arXiv:2406.05074},
  year={2024}
}

@article{zimmermann2024virchow2,
  title={Virchow2: Scaling self-supervised mixed magnification models in pathology},
  author={Zimmermann, Eric and Vorontsov, Eugene and Viret, Julian and Casson, Adam and Zelechowski, Michal and Shaikovski, George and Tenenholtz, Neil and Hall, James and Klimstra, David and Yousfi, Razik and others},
  journal={arXiv preprint arXiv:2408.00738},
  year={2024}
}

@article{filiot2024phikonv2,
  title={Phikon-v2, a large and public feature extractor for biomarker prediction},
  author={Filiot, Alexandre and Jacob, Paul and Mac Kain, Alice and Saillard, Charlie},
  journal={arXiv preprint arXiv:2409.09173},
  year={2024}
}

@article{xiang2025musk,
  title={A vision--language foundation model for precision oncology},
  author={Xiang, Jinxi and Wang, Xiyue and Zhang, Xiaoming and Xi, Yinghua and Eweje, Feyisope and Chen, Yijiang and Li, Yuchen and Bergstrom, Colin and Gopaulchan, Matthew and Kim, Ted and others},
  journal={Nature},
  volume={638},
  pages={769--778},
  year={2025}
}

@article{ding2025titan,
  title={A multimodal whole-slide foundation model for pathology},
  author={Ding, Tong and Wagner, Sophia J and Song, Andrew H and Chen, Richard J and Lu, Ming Y and Zhang, Andrew and Vaidya, Anurag J and Jaume, Guillaume and Shaban, Muhammad and Kim, Ahrong and others},
  journal=NATMED,
  volume={31},
  pages={3749--3761},
  year={2025}
}

@article{scalbert2026hoptimus1,
  title={Abstract {LB174}: {H-optimus-1}: A foundation model for computational histopathology},
  author={Scalbert, Marin and Saillard, Charlie and Peeters, Thomas and Gonzalez, Liam and Valter, Dasha and Llinares-L{\'o}pez, Felipe and Mariet, Zelda E and Jenatton, Rodolphe},
  journal={Cancer Research},
  volume={86},
  number={8\_Supplement},
  pages={LB174--LB174},
  year={2026}
}

@article{kapse2026genbiopathfm,
  title={{GenBio-PathFM}: A State-of-the-Art Foundation Model for Histopathology},
  author={Kapse, Saarthak and Ayg{\"u}n, Mehmet and Cole, Elijah and Lundberg, Emma and Song, Le and Xing, Eric P},
  journal={bioRxiv},
  year={2026}
}

@inproceedings{li2025advancing,
  title={Advancing multiple instance learning with continual learning for whole slide imaging},
  author={Li, Xianrui and Cui, Yufei and Li, Jun and Chan, Antoni B},
  booktitle=CVPR,
  pages={20800--20809},
  year={2025}
}

@article{qu2022weno,
  title={Bi-directional weakly supervised knowledge distillation for whole slide image classification},
  author={Qu, Linhao and Wang, Manning and Song, Zhijian and others},
  journal=NIPS,
  volume={35},
  pages={15368--15381},
  year={2022}
}

@inproceedings{tourniaire2021mixed,
  title={Attention-based multiple instance learning with mixed supervision on the {CAMELYON16} dataset},
  author={Tourniaire, Paul and Ilie, Marius and Hofman, Paul and Ayache, Nicholas and Delingette, Herve},
  booktitle={MICCAI Workshop on Computational Pathology},
  pages={216--226},
  year={2021}
}

@inproceedings{sadafi2023active,
  title={Active learning enhances classification of histopathology whole slide images with attention-based multiple instance learning},
  author={Sadafi, Ario and Navab, Nassir and Marr, Carsten},
  booktitle={IEEE International Symposium on Biomedical Imaging (ISBI)},
  pages={1--5},
  year={2023}
}

@inproceedings{bontempo2023dasmil,
  title={{DAS-MIL}: Distilling across scales for {MIL} classification of histological {WSIs}},
  author={Bontempo, Gianpaolo and Porrello, Angelo and Bolelli, Federico and Calderara, Simone and Ficarra, Elisa},
  booktitle=MICCAI,
  pages={248--258},
  year={2023}
}

@inproceedings{shin2026lrmil,
  title={{LRMIL}: Efficient Low-Resolution Multiple Instance Learning via High-Resolution Knowledge Distillation for Whole Slide Image Classification},
  author={Shin, Yonghan and Jeong, Won-Ki},
  booktitle=MICCAI,
  pages={289--298},
  year={2026}
}

@article{nechaev2025spider,
  title={{SPIDER}: a comprehensive multi-organ supervised pathology dataset and baseline models},
  author={Nechaev, Dmitry and Pchelnikov, Alexey and Ivanova, Ekaterina},
  journal={arXiv preprint arXiv:2503.02876},
  year={2025}
}

@article{nechaev2025histai,
  title={{HistAI}: an open-source, large-scale whole slide image dataset for computational pathology},
  author={Nechaev, Dmitry and Pchelnikov, Alexey and Ivanova, Ekaterina},
  journal={arXiv preprint arXiv:2505.12120},
  year={2025}
}

@article{asadi2024ocean,
  title={Machine learning-driven histotype diagnosis of ovarian carcinoma: insights from the {OCEAN} {AI} challenge},
  author={Asadi-Aghbolaghi, Maryam and Farahani, Hossein and Zhang, Allen and Akbari, Ardalan and Kim, Sirim and Chow, Ashley and Dane, Sohier and OCEAN Challenge Consortium and OTTA Consortium and Huntsman, David G and others},
  journal={medRxiv},
  year={2024}
}

@article{xu2021predicting,
  title={Predicting axillary lymph node metastasis in early breast cancer using deep learning on primary tumor biopsy slides},
  author={Xu, Feng and Zhu, Chuang and Tang, Wenqi and Wang, Ying and Zhang, Yu and Li, Jie and Jiang, Hongchuan and Shi, Zhongyue and Liu, Jun and Jin, Mulan},
  journal={Frontiers in Oncology},
  volume={11},
  pages={759007},
  year={2021}
}

@article{roetzer2022digital,
  title={The digital brain tumour atlas, an open histopathology resource},
  author={Roetzer-Pejrimovsky, Thomas and Moser, Anna-Christina and Atli, Baran and Vogel, Clemens Christian and Mercea, Petra A and Prihoda, Romana and Gelpi, Ellen and Haberler, Christine and H{\"o}ftberger, Romana and Hainfellner, Johannes A and others},
  journal={Scientific Data},
  volume={9},
  pages={55},
  year={2022}
}

@article{gillette2020proteogenomic,
  title={Proteogenomic characterization reveals therapeutic vulnerabilities in lung adenocarcinoma},
  author={Gillette, Michael A and Satpathy, Shankha and Cao, Song and Dhanasekaran, Saravana M and Vasaikar, Suhas V and Krug, Karsten and Petralia, Francesca and Li, Yize and Liang, Wen-Wei and Reva, Boris and others},
  journal={Cell},
  volume={182},
  pages={200--225},
  year={2020}
}

@article{satpathy2021proteogenomic,
  title={A proteogenomic portrait of lung squamous cell carcinoma},
  author={Satpathy, Shankha and Krug, Karsten and Beltran, Pierre M Jean and Savage, Sara R and Petralia, Francesca and Kumar-Sinha, Chandan and Dou, Yongchao and Reva, Boris and Kane, M Harry and Avanessian, Shayan C and others},
  journal={Cell},
  volume={184},
  pages={4348--4371},
  year={2021}
}

@article{clark2019integrated,
  title={Integrated proteogenomic characterization of clear cell renal cell carcinoma},
  author={Clark, David J and Dhanasekaran, Saravana M and Petralia, Francesca and Pan, Jianbo and Song, Xiaoyu and Hu, Yingwei and da Veiga Leprevost, Felipe and Reva, Boris and Lih, Tung-Shing M and Chang, Hui-Yin and others},
  journal={Cell},
  volume={179},
  pages={964--983},
  year={2019}
}

@article{zhang2025standardizing,
  title={Accelerating data processing and benchmarking of {AI} models for pathology},
  author={Zhang, Andrew and Jaume, Guillaume and Vaidya, Anurag and Ding, Tong and Mahmood, Faisal},
  journal={arXiv preprint arXiv:2502.06750},
  year={2025}
}

@inproceedings{kornblith2019similarity,
  title={Similarity of neural network representations revisited},
  author={Kornblith, Simon and Norouzi, Mohammad and Lee, Honglak and Hinton, Geoffrey},
  booktitle=ICML,
  pages={3519--3529},
  year={2019}
}

@article{achille2017critical,
  title={Critical learning periods in deep neural networks},
  author={Achille, Alessandro and Rovere, Matteo and Soatto, Stefano},
  journal={arXiv preprint arXiv:1711.08856},
  year={2017}
}

@article{golatkar2019time,
  title={Time matters in regularizing deep networks: Weight decay and data augmentation affect early learning dynamics, matter little near convergence},
  author={Golatkar, Aditya Sharad and Achille, Alessandro and Soatto, Stefano},
  journal=NIPS,
  volume={32},
  year={2019}
}

@inproceedings{xuhong2018explicit,
  title={Explicit inductive bias for transfer learning with convolutional networks},
  author={Xuhong, LI and Grandvalet, Yves and Davoine, Franck},
  booktitle=ICML,
  pages={2825--2834},
  year={2018}
}

@inproceedings{lin2014microsoft,
  title={Microsoft {COCO}: Common objects in context},
  author={Lin, Tsung-Yi and Maire, Michael and Belongie, Serge and Hays, James and Perona, Pietro and Ramanan, Deva and Doll{\'a}r, Piotr and Zitnick, C Lawrence},
  booktitle=ECCV,
  pages={740--755},
  year={2014}
}

@misc{loeffler2021tcga,
  title        = {Manual tumor annotations in {TCGA}},
  author       = {Loeffler, Chiara and Kather, Jakob Nikolas},
  howpublished = {Zenodo, doi:10.5281/zenodo.5320076},
  year         = {2021}
}
}

\providecommand{\tPHrow}{%
  & \tPH & \tPH & \tPH & \tPH & \tPH & \tPH & \tPH & \tPH & \tPH & \tPH
  & \tPH & \tPH & \tPH & \tPH & \tPH}
 
\clearpage
\maketitlesupplementary
\appendix
 
\setcounter{table}{0}
\setcounter{figure}{0}
\setcounter{equation}{0}
\renewcommand{\thetable}{S\arabic{table}}
\renewcommand{\theHtable}{S\arabic{table}}
\renewcommand{\thefigure}{S\arabic{figure}}
\renewcommand{\theHfigure}{S\arabic{figure}}
\renewcommand{\theequation}{S\arabic{equation}}
\renewcommand{\theHequation}{S\arabic{equation}}

\section{Method details}

\subsection{Anchor bank}
\label{app:anchor}

The bank holds 768 pathology patches, 256 from each of TCGA, HistAI, and SPIDER~\cite{nechaev2025spider}; none comes from a downstream dataset.
From TCGA and from HistAI, we select 128 pretraining slides at random, approximately balanced across diagnostic classes, and take two patches from each: one from the top 25\% and one from the 25--75\% range of attention under an earlier teacher consensus, so that the bank covers both highly and moderately attended tissue.
From SPIDER, we sample 256 crops at random, balanced across classes.
Class prototypes, a facility-location subset, and whitened similarities did not improve lesion AUROC over these anchors with cosine similarity (all within $\pm0.002$), so we use them without further optimization.

\subsection{Anchor-coordinate alignment: all PE pairs and anchor controls}
\label{app:cosine}

\begin{figure}[t]
\centering
\includegraphics[width=\linewidth]{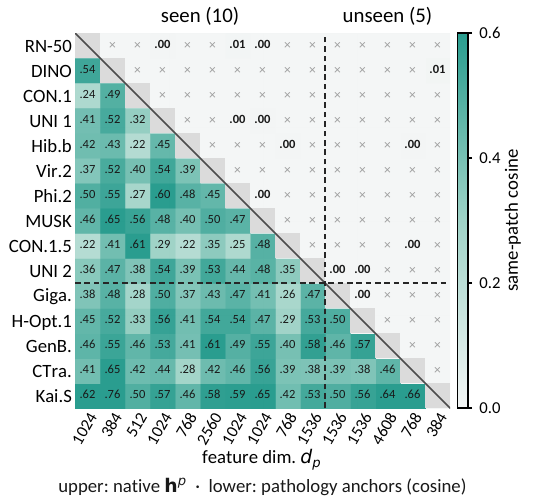}
\caption{\textbf{Same-patch cosine for all PE pairs.} As in main \cref{fig:anchor}A: test patches of \Dall{} (50 slides per dataset, up to 200 patches per slide), mean over datasets. Upper triangle: absolute cosine of native features; $\times$: dimensions differ. Lower triangle: anchor coordinates. Dashed lines: seen (10) vs.\ unseen (5) PEs.}
\label{fig:cosine_all}
\end{figure}

\Cref{fig:cosine_all} extends main \cref{fig:anchor}A to all 15 PEs.
Native features (upper triangle) are comparable only for the 13 pairs of equal width, all at $\approx$0; anchor coordinates (lower triangle) are comparable for all 105 pairs.

\paragraph{Anchor controls.}
Main \cref{fig:anchor}B replaces the pathology anchors with two controls of the same size.
The natural-image control uses 768 photographs drawn at random from COCO val2017~\cite{lin2014microsoft}, centre-cropped and resized to $256{\times}256$ pixels.
The uniform-noise control uses 768 images of the same size with i.i.d.\ uniform RGB values; every PE embeds them almost identically, so the coordinates concentrate on one direction (effective rank $1.05$), and a student pretrained on them falls to chance on unseen PEs (main \cref{tab:ablation}).

\subsection{Teacher training details}
\label{app:teacher}

Each teacher is a gated-attention ABMIL trained on the native features of its PE using slide-level diagnostic labels.
To encourage diverse class-relevant evidence selection, we supplement slide-level cross-entropy with the attention regularization of ACMIL~\cite{zhang2024attention} and AttriMIL~\cite{cai2025attrimil}.
The teacher objective is
\begin{equation}
\mathcal{L}_T = \mathcal{L}_{\mathrm{CE}} +
\mathcal{L}_{\mathrm{ACMIL}} + \mathcal{L}_{\mathrm{AttriMIL}}.
\label{eq:teacher}
\end{equation}
For ACMIL, we use five attention branches with
$\lambda_{\mathrm{br}}=\lambda_{\mathrm{div}}=1$
for branch-level cross-entropy and diversity regularization.
For AttriMIL, we set $\lambda_{\mathrm{evi}}=2$ and
$\lambda_{\mathrm{sp}}=0.1$ for the attribute and spatial constraints, respectively.
After pretraining, all teachers are frozen and used solely to provide attention targets $\boldsymbol{\alpha}^{t}$ for student training.

\subsection{Teacher PE selection}
\label{app:teacher_selection}

Teachers are selected by classification accuracy on the pretraining validation set; no downstream data is involved.
Among the ten seen PEs, ResNet-50 (40.7\%) and Lunit DINO ViT-S/16 (70.9\%) achieve the two lowest accuracies (ten-PE mean: 71.6\%) and are excluded as teachers, while remaining student inputs.
Including them does not improve the consensus: its attention precision on the annotated TCGA validation slides (\cref{app:impl}) is 0.903 with all ten teachers and 0.905 with the selected eight.

\subsection{Student objectives}
\label{app:student}
$\mathcal{L}_{\mathrm{inv}}$ in main \cref{eq:student} aligns two views $v,u$ of the same slide, read through two different seen PEs, on their shared patches:
\begin{equation}
\begin{split}
\mathcal{L}_{\mathrm{inv}}=\;&\lambda_1\,\mathcal{L}_{\mathrm{NCE}}(\mathbf{e}^{v},\mathbf{e}^{u})+\lambda_2\,\tfrac{1}{N'}\|\tilde{\boldsymbol{\ell}}^{v}-\tilde{\boldsymbol{\ell}}^{u}\|_2^2\\
&+\lambda_3\,\big(1-\cos(\mathbf{s}^{v},\mathbf{s}^{u})\big),
\end{split}
\label{eq:supp_inv}
\end{equation}
where $\mathcal{L}_{\mathrm{NCE}}$ is a symmetric InfoNCE loss between the student's patch embeddings $\mathbf{e}$ over up to 256 shared patches (temperature 0.1; the same patch under the other PE is the positive), $\tilde{\boldsymbol{\ell}}$ are the standardized attention scores of the $N'$ shared patches, and $\mathbf{s}$ the slide embeddings.
We use $\lambda_1=\lambda_2=\lambda_3=0.5$, and $\gamma=1$ in main \cref{eq:student}.

\section{Experimental details}

\subsection{Pretraining data}
\label{app:pretraining}

The teachers and student are pretrained on 19{,}766 labeled
WSIs from TCGA~\cite{weinstein2013cancer} (10{,}937)
and HistAI~\cite{nechaev2025histai} (8{,}829),
covering 62 diagnostic classes across 23 organs.
Classes are defined from dataset diagnostic labels,
retaining those with at least 20 training patients.
Training and validation sets are split at the patient
or case level: 9{,}872/1{,}065 for TCGA and
7{,}862/967 for HistAI.
Of the TCGA validation slides, 925 have public manual
tumor annotations~\cite{loeffler2021tcga}, which we use
for checkpoint selection (\cref{app:impl}).
HistAI additionally provides benign and premalignant
diagnoses absent from TCGA.
We checked slide filenames and patient/case identifiers
between the source datasets (11{,}339 TCGA and
14{,}637 HistAI WSIs) and all eight downstream
benchmarks, finding no exact matches.

\subsection{Patch encoders}
\label{app:encoder}

All patch features are extracted from non-overlapping
$256\times256$ tissue patches using
TRIDENT~\cite{zhang2025standardizing}.
\Cref{tab:pe_list} lists the 15 PEs: ten \emph{seen}
PEs used during student pretraining, including eight
teacher PEs, and five \emph{unseen} PEs used only
downstream.
Supporting an unseen PE requires only a one-time
encoding of the anchor bank.

\begin{table}[t]
\centering
\caption{
\textbf{The 15 patch encoders.}
Dim: feature dimension; T: teacher PE.
}
\label{tab:pe_list}
\footnotesize
\setlength{\tabcolsep}{3pt}
\renewcommand{\arraystretch}{0.95}
\begin{tabular*}{\columnwidth}{@{\extracolsep{\fill}}lrc@{}}
\toprule
PE & Dim & T \\
\midrule
\multicolumn{3}{@{}l}{\textit{Seen PEs (10)}} \\
ResNet-50~\cite{he2016deep}
    & 1024 & -- \\
DINO (Lunit)~\cite{caron2021emerging,kang2023benchmarking}
    & 384 & -- \\
CONCH~\cite{lu2024visual}
    & 512 & $\checkmark$ \\
UNI~\cite{chen2024uni}
    & 1024 & $\checkmark$ \\
Hibou-B~\cite{nechaev2024hibou}
    & 768 & $\checkmark$ \\
Virchow2~\cite{zimmermann2024virchow2}
    & 2560 & $\checkmark$ \\
Phikon-v2~\cite{filiot2024phikonv2}
    & 1024 & $\checkmark$ \\
MUSK~\cite{xiang2025musk}
    & 1024 & $\checkmark$ \\
CONCH v1.5~\cite{ding2025titan}
    & 768 & $\checkmark$ \\
UNI-2~\cite{chen2024uni}
    & 1536 & $\checkmark$ \\
\midrule
\multicolumn{3}{@{}l}{\textit{Unseen PEs (5)}} \\
CTransPath~\cite{wang2022ctranspath}
    & 768 & -- \\
Kaiko-S~\cite{ai2024largescale}
    & 384 & -- \\
GigaPath~\cite{xu2024gigapath}
    & 1536 & -- \\
H-Optimus-1~\cite{scalbert2026hoptimus1}
    & 1536 & -- \\
GenBio-PathFM~\cite{kapse2026genbiopathfm}
    & 4608 & -- \\
\bottomrule
\end{tabular*}
\end{table}

\subsection{Downstream benchmarks and annotations}
\label{app:downstream}
\label{app:annotations}

\Cref{tab:data} summarizes the twelve tasks across
eight benchmarks (\Dall).
BRACS and BCNB use official splits.
CAMELYON16 and EBRAINS retain their designated
test sets and hold out training data for validation.
UBC-OCEAN and PANDA use stratified 60/20/20
splits of public training data; PANDA additionally
groups serial sections by biopsy and stratifies by scanner.
CPTAC follows five patient-level splits from
Patho-Bench~\cite{zhang2025standardizing}.
Results are averaged over five seeds, with multiple
tasks averaged within each benchmark.
For the four annotated datasets (\Dann), region
annotations are projected onto the shared patch grid.
Patches with at least 50\% lesion coverage are
labeled positive, those with no lesion coverage
negative, and intermediate patches are excluded.
In main \cref{fig:entangle}D and main \cref{fig:lost}, a model
is said to locate the lesion of a slide when its lesion
AUROC on that slide is at least $\tau{=}0.75$; the same
threshold is applied to the prior before downstream
training and to the model after it.
Annotations are used only for attention evaluation,
never for training.
Experiments on \Dann{} that train a downstream model
use the 7-class task for BRACS.

All slides used for pretraining, the anchor bank, and
downstream evaluation come from public, de-identified
datasets, obtained through each provider's access
procedure and used under its data-use terms; no new
patient data were collected.

\begin{table}[t]
\centering
\caption{
\textbf{Downstream benchmarks (\Dall).}
Slides: train/validation/test; CPTAC reports total
slides across five patient-level splits.
$^{\dagger}$Datasets with region annotations (\Dann).
C: number of classes;
bAcc: balanced accuracy;
Acc: accuracy;
QWK: quadratic weighted kappa.
}
\label{tab:data}
\footnotesize
\setlength{\tabcolsep}{3pt}
\renewcommand{\arraystretch}{1.08}
\resizebox{\columnwidth}{!}{%
\begin{tabular}{@{}l l c l c@{}}
\toprule
Dataset & Task & C & Slides & Metric \\
\midrule
BRACS$^{\dagger}$~\cite{brancati2022bracs}
    & 3-class subtyping & 3
    & 395/65/87 & bAcc \\
    & 7-class subtyping & 7
    & 395/65/87 & bAcc \\

CAMELYON16$^{\dagger}$~\cite{ehteshami2017diagnostic}
    & Metastasis & 2
    & 216/54/129 & Acc \\

UBC-OCEAN$^{\dagger}$~\cite{asadi2024ocean}
    & Subtyping & 5
    & 307/103/103 & bAcc \\

PANDA$^{\dagger}$~\cite{bulten2022artificial}
    & ISUP grading & 6
    & 6360/2130/2125 & QWK \\

BCNB~\cite{xu2021predicting}
    & ER & 2
    & 630/210/218 & Acc \\
    & PR & 2
    & 630/210/218 & Acc \\
    & HER2 & 2
    & 630/210/218 & Acc \\

CPTAC~\cite{gillette2020proteogenomic,satpathy2021proteogenomic}
    & NSCLC subtyping & 2
    & 309 (total) & Acc \\

CPTAC~\cite{clark2019integrated}
    & CCRCC grading & 2
    & 524 (total) & Acc \\

EBRAINS~\cite{roetzer2022digital}
    & 12-class subtyping & 12
    & 1159/587/573 & bAcc \\
    & 30-class subtyping & 30
    & 1159/587/573 & bAcc \\
\bottomrule
\end{tabular}%
}
\end{table}

\subsection{Implementation}
\label{app:impl}

Unless otherwise specified, we use gated-attention
ABMIL~\cite{ilse2018attention,shao2025transfer}
as the default architecture for the teachers, student,
and downstream aggregators, with 512-d embeddings
and 384-d attention.
We choose ABMIL for two reasons: it achieves the highest
average performance after pretraining among 11 MIL
architectures~\cite{shao2025transfer}, and its explicit
patch-level attention enables direct distillation
of the pretrained prior without architectural modifications.
Scratch baselines use the same downstream architecture
and training protocol without AKD.
Pretrained baselines (\cref{app:baselines}) are run with
their released weights on features from our pipeline
(\cref{app:encoder}), so that all methods share the same
inputs.

\begin{figure}[t]
\centering
\includegraphics[width=\linewidth]{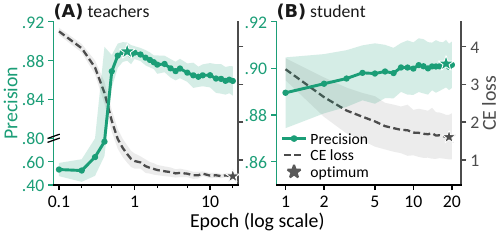}
\caption{\textbf{Checkpoint selection.} Attention precision (tumor fraction among the 5\% most attended patches) on the annotated TCGA validation slides, and cross-entropy on the full validation set (TCGA and HistAI), during pretraining. (A) Teachers, trained beyond their stopping point for illustration. (B) Student. Lines: mean over the 10 seen PEs; bands: s.d.\ across PEs. Stars: best value of each curve.}
\label{fig:supp_ckpt}
\end{figure}

All models use AdamW.
Teachers use a constant learning rate of $10^{-4}$; the
student is trained for 20 epochs with $2\times10^{-4}$ and
cosine decay; both use weight decay 0.04.
As the prior supplies attention, pretraining checkpoints are
selected by attention precision on the annotated TCGA
validation slides (\cref{app:pretraining}): the tumor
fraction among the 5\% most attended patches (3\% and 10\%
give nearly identical results). R-precision is uninformative
here, as the coarse outlines label about half of each slide
as tumor.
These annotations enter no training loss; they are used only
to select checkpoints, and the annotated slides are disjoint
from all downstream benchmarks.
Trained on slide labels alone, a teacher's attention
precision peaks near epoch 1 and then declines while its
cross-entropy keeps decreasing (\cref{fig:supp_ckpt}A).
Teachers are therefore evaluated every 0.1 epoch from
epoch 0.6 and stopped after five evaluations without
improvement; the student shows no such decline
(\cref{fig:supp_ckpt}B).
Downstream aggregators use $E=20$ epochs,
learning rate $10^{-4}$ with cosine decay, weight decay
$10^{-5}$, and $\beta_0=0.3$, decayed linearly to zero over
the first 10 epochs (main \cref{eq:beta}); we report the
checkpoint with the best validation score.
For $K$-shot experiments, $K\in\{4,8,16,32\}$ slides per
class are sampled using nested, seed-matched draws shared
across PEs and methods, with epochs and the distillation
schedule scaled to match the full-data training updates.

\subsection{Compute}
\label{app:compute}

\paragraph{Software and hardware.}
Features were extracted with TRIDENT using PyTorch 2.8.0.
Experiments ran on single GPUs (NVIDIA RTX A6000, RTX 3080 Ti, and RTX 2080 Ti); all timings below use one RTX A6000 (48\,GB).

\paragraph{Computational cost.}
\Cref{tab:cost} compares ours with scratch ABMIL
across the feature dimensions of the 15 PEs.
Our method precomputes and caches attention targets
from \scout before downstream training.
Timings are measured in fp32 on BRACS with features held
in GPU memory and reported as the median of three runs.
Including caching, our method requires only
$1.05$--$1.10\times$ the training time of scratch
ABMIL, with no additional inference cost.

\begin{table}[t]
\centering
\caption{
\textbf{Computational cost relative to scratch ABMIL.}
$d$: PE feature dimension.
Training: wall-clock time for 20 epochs;
Cache: one-time attention-target computation;
Ours: training time including the cache;
$\times$: ours relative to scratch ABMIL.
Inference: forward time for a slide with
$N=5{,}000$ patches.
}
\label{tab:cost}
\footnotesize
\setlength{\tabcolsep}{3pt}
\renewcommand{\arraystretch}{1.1}
\begin{tabular*}{\columnwidth}{@{\extracolsep{\fill}} l cccc ccc @{}}
\toprule
& \multicolumn{4}{c}{Training, 20 epochs (s)}
& \multicolumn{3}{c}{Inference, per slide (ms)} \\
\cmidrule(lr){2-5}\cmidrule(lr){6-8}
$d$ & Scratch & Cache & Ours & $\times$
    & Scratch & Ours & $\times$ \\
\midrule
384     & 15.5 & 0.65 & 17.0 & 1.10 & 0.62 & 0.62 & 1.00 \\
512     & 15.9 & 0.69 & 17.2 & 1.08 & 0.65 & 0.65 & 1.00 \\
768     & 17.4 & 0.79 & 18.7 & 1.07 & 0.71 & 0.71 & 1.00 \\
1{,}024 & 19.4 & 0.89 & 20.7 & 1.06 & 0.77 & 0.77 & 1.00 \\
1{,}536 & 22.2 & 1.15 & 23.8 & 1.07 & 0.90 & 0.90 & 1.00 \\
2{,}560 & 27.7 & 1.10 & 29.0 & 1.05 & 1.16 & 1.16 & 1.00 \\
4{,}608 & 36.6 & 2.36 & 39.6 & 1.08 & 1.69 & 1.69 & 1.00 \\
\bottomrule
\end{tabular*}
\end{table}

\section{Additional results}

\subsection{Localization per dataset}
\label{app:prior_per_dataset}

\Cref{tab:prior_per_dataset} reports zero-shot lesion AUROC and lesion R-precision on
the test slides of \Dann, separately averaged
over seen and unseen PEs.
\scout achieves higher lesion
AUROC than task-trained scratch ABMIL on all four
datasets when averaged over 15 PEs.
Taken together, without any task label the prior exceeds
task-trained ABMIL in lesion AUROC on every dataset for
seen and unseen PEs alike, except for a 0.002 deficit on
CAMELYON16 with unseen PEs.
In R-precision the two are on par on average, and scratch
is ahead only on CAMELYON16, where the task label is lesion
presence itself.

\begin{table}[t]
\centering
\caption{
\textbf{Zero-shot localization per dataset.}
Lesion AUROC and lesion R-precision of
\scout and task-trained scratch ABMIL.
Seen and unseen results are averaged over
10 and 5 PEs, respectively.
}
\label{tab:prior_per_dataset}
\footnotesize
\setlength{\tabcolsep}{3pt}
\renewcommand{\arraystretch}{1.08}
\begin{tabular*}{\columnwidth}{@{\extracolsep{\fill}}llcccc@{}}
\toprule
& & \multicolumn{2}{c}{AUROC}
& \multicolumn{2}{c}{R-precision} \\
\cmidrule(lr){3-4}\cmidrule(lr){5-6}
Dataset & PEs & \scout & Scratch
& \scout & Scratch \\
\midrule
CAMELYON16 & Seen   & 0.981 & 0.968 & 0.769 & 0.850 \\
           & Unseen & 0.975 & 0.977 & 0.765 & 0.877 \\
BRACS      & Seen   & 0.945 & 0.879 & 0.180 & 0.168 \\
           & Unseen & 0.940 & 0.847 & 0.177 & 0.132 \\
UBC-OCEAN  & Seen   & 0.861 & 0.811 & 0.583 & 0.506 \\
           & Unseen & 0.847 & 0.815 & 0.560 & 0.510 \\
PANDA      & Seen   & 0.934 & 0.889 & 0.727 & 0.709 \\
           & Unseen & 0.933 & 0.894 & 0.728 & 0.725 \\
\midrule
Mean       & Seen   & 0.930 & 0.886 & 0.565 & 0.558 \\
           & Unseen & 0.924 & 0.883 & 0.558 & 0.561 \\
\bottomrule
\end{tabular*}
\end{table}

\subsection{Attention consistency across PEs}
\label{app:consistency}

\begin{figure}[t]
\centering
\includegraphics[width=\linewidth]{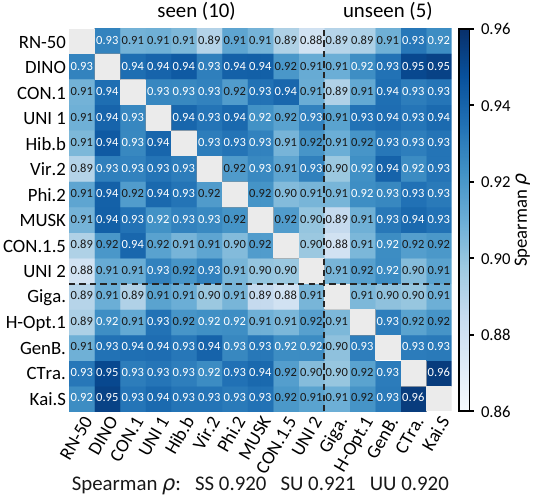}
\caption{\textbf{Attention consistency across PEs.} Mean within-slide Spearman correlation of \scout's patch attention between two PEs, over the test slides of \Dann.}
\label{fig:consistency}
\end{figure}

For each test slide of \Dann, we compute the
Spearman rank correlation between
\scout's patch attention under every pair of PEs
and average over slides (\cref{fig:consistency}).
Across all 105 PE pairs, the mean correlation is
$0.92$, with seen--seen, seen--unseen, and
unseen--unseen pairs all yielding means between
$0.920$ and $0.921$.
This consistency extends to PEs excluded from
pretraining.

\subsection{Zero-shot localization baselines}
\label{app:baselines}

\paragraph{Zero-shot patch scoring.}\Cref{tab:zs_baselines} summarizes the nine baselines
in main \cref{fig:prior}.
All methods score patches on the same spatial grid
without using slides or lesion annotations from
\Dann{} for training.
Each baseline is compared with \scout
on its native PE, except SPIDER LP, which is
evaluated across all 15 PEs.

\begin{table}[t]
\centering
\caption{
\textbf{Zero-shot patch-scoring baselines.}
Each method produces patch-level scores without
target-task training.
Prompt (2 VLMs) denotes the mean over CONCH and MUSK;
SPIDER LP is trained separately for each PE.
}
\label{tab:zs_baselines}
\footnotesize
\setlength{\tabcolsep}{2.5pt}
\renewcommand{\arraystretch}{1.08}
\begin{tabular*}{\columnwidth}{@{\extracolsep{\fill}}l l p{0.47\columnwidth}@{}}
\toprule
Method & Native PE & Patch score \\
\midrule
Shao et al.\ U~\cite{shao2025transfer}
& UNI-2
& Pretrained ABMIL attention \\

Shao et al.\ C~\cite{shao2025transfer}
& CONCH v1.5
& Pretrained ABMIL attention \\

Fu et al.~\cite{fu2026pretraining}
& CONCH v1.5
& Pretrained V-ABMIL attention \\

COBRA~\cite{lenz2025cobra}
& Virchow2
& Patch-level softmax attention \\

TITAN~\cite{ding2025titan}
& CONCH v1.5
& Last-block class-token attention \\

PRISM2~\cite{vorontsov2026end}
& Virchow2
& Pooled Perceiver query attention \\

CARE~\cite{zhang2026care}
& CONCH v1.5
& Region and within-region attention \\

\begin{tabular}[t]{@{}l@{}}Prompt (2 VLMs)\\\cite{lu2024visual,xiang2025musk}\end{tabular}
& CONCH / MUSK
& Lesion-minus-normal text\newline similarity \\

SPIDER LP~\cite{nechaev2025spider}
& All 15 PEs
& Linear-probe tumor probability \\
\bottomrule
\end{tabular*}
\end{table}

For the prompt baseline, CONCH and MUSK use
22 text templates (e.g., ``an H\&E image of
\textsc{classname}.'').
Each class embedding is the normalized mean of
its template embeddings.
Patch scores are computed as the maximum cosine
similarity to lesion prompts minus that to normal
prompts, using each model's joint vision--language
space.
The lesion and normal classes are defined by
dataset diagnosis: metastatic carcinoma versus
normal lymph node in CAMELYON16; benign and
malignant breast lesions versus normal breast
in BRACS; five ovarian carcinoma subtypes versus
normal stroma in UBC-OCEAN; and prostate carcinoma
versus benign glands in PANDA.

For SPIDER LP, we train a separate class-balanced
logistic regression for each of the 15 PEs using
L2-normalized SPIDER patch features.
The training set contains 10{,}350 patches from
69 classes across four organs; 17 malignant
classes are labeled as tumor, and the others
as non-tumor.
Regularization is selected by five-fold
slide-grouped cross-validation.
SPIDER contains no ovarian or prostate tissue.

\paragraph{Downstream comparison.}
The released checkpoints of Shao
\etal~\cite{shao2025transfer} (UNI-2 and CONCH v1.5)
and Fu \etal~\cite{fu2026pretraining}
(CONCH v1.5) are fine-tuned in their original
aggregator architectures, with classifiers
initialized anew.
For non-native PEs, trainable linear adapters
map patch features to the checkpoint's input
dimension.

Slide foundation models
(COBRA~\cite{lenz2025cobra},
TITAN~\cite{ding2025titan},
PRISM2~\cite{vorontsov2026end},
CARE~\cite{zhang2026care})
use frozen slide encoders with linear probing.
For non-native PEs, a trainable linear adapter
is optimized jointly with the classifier.
All classifiers use the same 20-epoch AdamW
training protocol.

Additional MIL architectures
(DTFD-MIL~\cite{zhang2022dtfd},
TransMIL~\cite{shao2021transmil},
WiKG~\cite{li2024dynamic},
ILRA~\cite{xiang2023ilra})
use the same downstream training protocol
as scratch ABMIL, with AKD
applied to their corresponding patch-weighting
scores.

\subsection{Qualitative examples}
\label{app:qualitative}

\begin{figure*}[t]
\centering
\includegraphics[width=\linewidth]{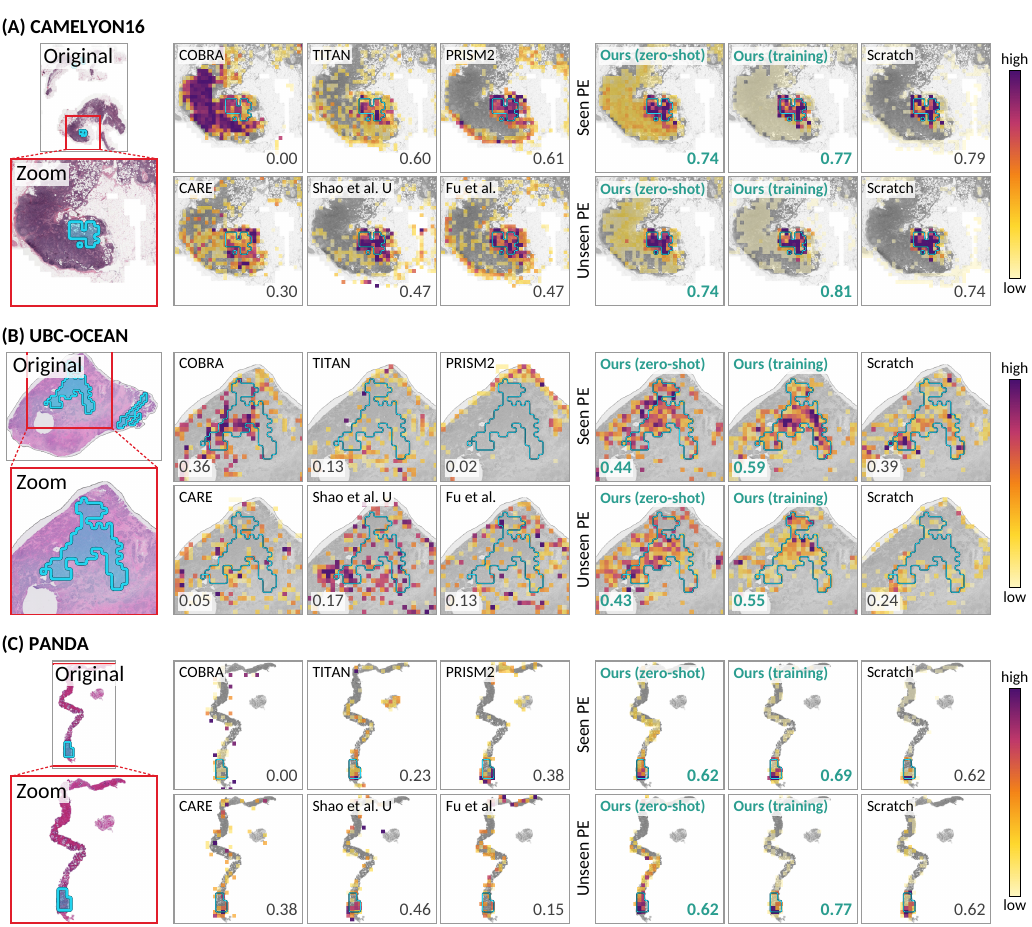}
\caption{
\textbf{Patch attention on slides from the other three annotated datasets.}
\textbf{(A)} CAMELYON16, \textbf{(B)} UBC-OCEAN, \textbf{(C)} PANDA.
Cyan: lesion annotation. Left: existing pretrained models on their native PEs. Right: \scout (Ours, zero-shot), the aggregator trained with it (Ours, training), and scratch ABMIL under a seen PE (UNI-2) and an unseen PE (GigaPath).
Numbers: lesion R-precision on the slide.
}
\label{fig:qual_appendix}
\end{figure*}

\Cref{fig:qual_appendix} shows patch attention on one slide each from CAMELYON16, UBC-OCEAN, and PANDA, in the layout of main \cref{fig:qual}.
On all three slides, \scout locates the annotated region zero-shot, with nearly identical lesion R-precision under the seen and the unseen PE.
The aggregator trained with it suppresses attention outside the annotation and reaches a higher R-precision in every case, refining the prior rather than merely copying it.
Existing pretrained models, in contrast, mostly fall below scratch ABMIL, consistent with main \cref{fig:prior}.
On CAMELYON16, where the task label is lesion presence itself, scratch ABMIL also localizes the metastasis (\cref{tab:prior_per_dataset}), and the aggregator trained with the prior keeps pace with it.

\begin{table*}[!b]
\centering
\caption{
\textbf{BRACS 3-class subtyping across 15 PEs.}
Balanced accuracy (\%), s.d.\ in parentheses.
Gray: native PE(s) of each method (ours: the 10 seen PEs).
Green/red: above/below scratch ABMIL.
Bold: best per PE.
}
\label{tab:bracs3}

\scriptsize
\setlength{\tabcolsep}{2.4pt}
\renewcommand{\arraystretch}{1.07}

\resizebox{\textwidth}{!}{%
\begin{tabular}{
    l
    *{10}{c}
    @{\hspace{3pt}}
    *{5}{c}
}

\toprule

& \multicolumn{10}{c}{Seen during our pretraining}
& \multicolumn{5}{c}{Unseen} \\

\cmidrule(lr){2-11}
\cmidrule(lr){12-16}

Method
& \thead{RN-50\\\cite{he2016deep}}
& \thead{DINO\\\cite{caron2021emerging}}
& \thead{CONCH\\\cite{lu2024visual}}
& \thead{UNI\\\cite{chen2024uni}}
& \thead{Hibou-B\\\cite{nechaev2024hibou}}
& \thead{Virch2\\\cite{zimmermann2024virchow2}}
& \thead{Phikon2\\\cite{filiot2024phikonv2}}
& \thead{MUSK\\\cite{xiang2025musk}}
& \thead{CON1.5\\\cite{ding2025titan}}
& \thead{UNI2\\\cite{chen2024uni}}
& \thead{CTrans\\\cite{wang2022ctranspath}}
& \thead{Kaiko-S\\\cite{ai2024largescale}}
& \thead{GigaP\\\cite{xu2024gigapath}}
& \thead{H-Opt1\\\cite{scalbert2026hoptimus1}}
& \thead{GenBio\\\cite{kapse2026genbiopathfm}} \\

\midrule

\multicolumn{16}{l}{\textit{Baseline MIL}} \\

DTFD-MIL~\cite{zhang2022dtfd}
& \mstd{59.9}{2.5}
& \mstd{68.8}{2.0}
& \mstd{72.1}{5.1}
& \mstd{69.6}{6.8}
& \mstd{67.7}{5.2}
& \mstd{74.9}{4.3}
& \mstd{64.9}{3.6}
& \mstd{68.4}{2.6}
& \mstd{75.9}{4.4}
& \mstd{70.9}{3.9}
& \mstd{64.5}{3.7}
& \mstd{67.4}{4.3}
& \mstd{68.7}{3.5}
& \mstd{68.4}{5.3}
& \mstd{70.1}{2.4}
\\

TransMIL~\cite{shao2021transmil}
& \mstd{53.0}{5.3}
& \mstd{64.1}{4.8}
& \mstd{70.0}{3.9}
& \mstd{62.5}{3.8}
& \mstd{60.7}{4.7}
& \mstd{63.7}{5.2}
& \mstd{58.6}{5.0}
& \mstd{64.9}{4.7}
& \mstd{71.3}{3.0}
& \mstd{60.2}{4.7}
& \mstd{59.6}{3.7}
& \mstd{61.5}{3.2}
& \mstd{62.0}{3.3}
& \mstd{60.2}{5.3}
& \mstd{69.2}{3.1}
\\

WiKG~\cite{li2024dynamic}
& \mstd{57.8}{4.7}
& \mstd{56.6}{6.7}
& \mstd{70.8}{5.0}
& \mstd{64.9}{4.8}
& \mstd{59.5}{4.9}
& \mstd{72.4}{3.4}
& \mstd{62.9}{6.5}
& \mstd{63.2}{5.8}
& \mstd{70.5}{4.6}
& \mstd{68.7}{5.5}
& \mstd{64.7}{3.9}
& \mstd{60.7}{8.6}
& \mstd{69.5}{5.5}
& \mstd{71.5}{3.8}
& \mstd{73.2}{6.2}
\\

ILRA~\cite{xiang2023ilra}
& \mstd{59.4}{2.3}
& \mstd{65.9}{4.3}
& \mstd{71.9}{2.8}
& \mstd{66.3}{5.3}
& \mstd{66.2}{4.6}
& \mstd{67.6}{5.4}
& \mstd{60.6}{4.0}
& \mstd{68.2}{3.5}
& \mstd{76.8}{5.5}
& \mstd{65.4}{6.2}
& \mstd{65.0}{5.4}
& \mstd{65.5}{5.8}
& \mstd{63.3}{4.1}
& \mstd{66.1}{4.0}
& \mstd{73.1}{5.8}
\\

\midrule

\multicolumn{16}{l}{\textit{Slide Foundation Models}} \\

COBRA~\cite{lenz2025cobra}
& \mstd{60.1}{1.8}
& \mstd{\textbf{76.4}}{1.6}
& \mstd{\textbf{78.5}}{1.1}
& \tnat{\mstd{55.1}{3.6}}
& \mstd{67.0}{0.0}
& \tnat{\mstd{62.0}{0.4}}
& \mstd{69.7}{1.9}
& \mstd{\textbf{74.6}}{3.4}
& \mstd{74.6}{3.6}
& \mstd{67.9}{1.1}
& \tnat{\mstd{56.7}{0.3}}
& \mstd{73.8}{2.9}
& \mstd{71.3}{0.1}
& \mstd{70.2}{2.5}
& \mstd{73.2}{1.8}
\\

TITAN~\cite{ding2025titan}
& \mstd{60.8}{0.2}
& \mstd{68.5}{4.4}
& \mstd{74.0}{2.6}
& \mstd{73.9}{1.9}
& \mstd{68.5}{1.8}
& \mstd{68.1}{3.6}
& \mstd{68.2}{1.7}
& \mstd{67.3}{2.0}
& \tnat{\mstd{71.0}{1.3}}
& \mstd{72.4}{3.3}
& \mstd{60.2}{0.4}
& \mstd{\textbf{74.3}}{0.5}
& \mstd{70.4}{1.4}
& \mstd{\textbf{78.4}}{3.2}
& \mstd{66.8}{0.2}
\\

PRISM2~\cite{vorontsov2026end}
& \mstd{59.3}{1.0}
& \mstd{67.7}{0.4}
& \mstd{68.3}{0.4}
& \mstd{71.1}{0.2}
& \mstd{62.1}{2.2}
& \tnat{\mstd{75.7}{3.0}}
& \mstd{61.4}{6.7}
& \mstd{71.2}{7.2}
& \mstd{76.3}{6.4}
& \mstd{65.4}{1.9}
& \mstd{\textbf{72.7}}{6.0}
& \mstd{66.5}{1.8}
& \mstd{67.1}{11.6}
& \mstd{62.9}{4.2}
& \mstd{73.0}{4.4}
\\

CARE~\cite{zhang2026care}
& \mstd{60.2}{3.0}
& \mstd{68.2}{5.3}
& \mstd{71.1}{4.0}
& \mstd{69.2}{7.2}
& \mstd{66.9}{5.4}
& \mstd{70.1}{2.6}
& \mstd{65.1}{3.8}
& \mstd{66.0}{4.0}
& \tnat{\mstd{69.5}{1.6}}
& \mstd{73.7}{2.8}
& \mstd{61.6}{4.4}
& \mstd{70.3}{7.9}
& \mstd{\textbf{73.9}}{0.6}
& \mstd{74.9}{0.6}
& \mstd{74.9}{6.8}
\\

\midrule

\multicolumn{16}{l}{
    \textit{Pretrained Models (ABMIL)~\cite{ilse2018attention}}
} \\

Scratch
& \mstd{62.0}{2.1}
& \mstd{65.6}{4.9}
& \mstd{75.2}{4.5}
& \mstd{65.6}{4.7}
& \mstd{70.3}{3.9}
& \mstd{69.8}{5.6}
& \mstd{66.0}{5.2}
& \mstd{71.4}{4.4}
& \mstd{75.9}{3.6}
& \mstd{70.8}{4.8}
& \mstd{67.5}{4.2}
& \mstd{66.5}{4.2}
& \mstd{70.2}{4.9}
& \mstd{69.7}{4.5}
& \mstd{72.7}{4.2}
\\

Shao et al.\ U~\cite{shao2025transfer}
& \dn{\mstd{60.2}{3.2}}
& \up{\mstd{71.4}{3.0}}
& \dn{\mstd{74.5}{4.3}}
& \up{\mstd{67.0}{5.1}}
& \dn{\mstd{68.9}{3.4}}
& \up{\mstd{72.0}{2.4}}
& \dn{\mstd{65.6}{3.5}}
& \dn{\mstd{68.7}{2.8}}
& \dn{\mstd{72.1}{2.9}}
& \tnat{\up{\mstd{74.3}{2.8}}}
& \dn{\mstd{57.0}{3.7}}
& \up{\mstd{69.9}{3.2}}
& \dn{\mstd{69.4}{4.4}}
& \dn{\mstd{68.7}{3.6}}
& \dn{\mstd{69.2}{3.9}}
\\

Shao et al.\ C~\cite{shao2025transfer}
& \dn{\mstd{60.0}{1.5}}
& \up{\mstd{73.0}{3.5}}
& \up{\mstd{76.7}{3.2}}
& \up{\mstd{69.3}{4.4}}
& \dn{\mstd{69.8}{2.5}}
& \up{\mstd{70.0}{3.0}}
& \dn{\mstd{61.5}{5.2}}
& \dn{\mstd{68.6}{1.9}}
& \tnat{\up{\mstd{\textbf{80.5}}{3.8}}}
& \up{\mstd{72.2}{4.9}}
& \dn{\mstd{59.0}{3.3}}
& \up{\mstd{70.5}{5.3}}
& \dn{\mstd{66.1}{6.1}}
& \up{\mstd{70.9}{3.8}}
& \up{\mstd{73.7}{4.1}}
\\

Fu et al.~\cite{fu2026pretraining}
& \dn{\mstd{60.1}{3.9}}
& \dn{\mstd{57.4}{13.0}}
& \dn{\mstd{69.4}{11.8}}
& \dn{\mstd{57.9}{11.7}}
& \dn{\mstd{53.9}{11.5}}
& \up{\mstd{71.7}{5.8}}
& \dn{\mstd{64.2}{2.7}}
& \dn{\mstd{68.5}{3.9}}
& \tnat{\dn{\mstd{68.2}{10.8}}}
& \up{\mstd{72.7}{5.4}}
& \dn{\mstd{58.2}{3.8}}
& \dn{\mstd{65.3}{5.7}}
& \dn{\mstd{63.3}{6.5}}
& \dn{\mstd{62.1}{5.0}}
& \up{\mstd{72.9}{4.9}}
\\

\textbf{Ours}
& \tnat{\up{\mstd{\textbf{63.7}}{1.9}}}
& \tnat{\up{\mstd{73.4}{1.9}}}
& \tnat{\up{\mstd{77.7}{3.7}}}
& \tnat{\up{\mstd{\textbf{74.3}}{3.8}}}
& \tnat{\up{\mstd{\textbf{71.3}}{1.9}}}
& \tnat{\up{\mstd{\textbf{77.4}}{1.6}}}
& \tnat{\up{\mstd{\textbf{70.3}}{2.4}}}
& \tnat{\dn{\mstd{69.4}{3.2}}}
& \tnat{\up{\mstd{76.0}{3.5}}}
& \tnat{\up{\mstd{\textbf{75.5}}{1.0}}}
& \up{\mstd{67.8}{1.3}}
& \up{\mstd{71.3}{4.5}}
& \up{\mstd{73.2}{2.4}}
& \up{\mstd{72.9}{2.6}}
& \up{\mstd{\textbf{75.7}}{1.6}}
\\

\bottomrule
\end{tabular}%
}

\end{table*}

\subsection{Ablation details}
\label{app:ablation}

\paragraph{Prior delivery methods
(main \cref{tab:delivery}).}
All variants use the same \scout and the downstream protocol of \cref{app:impl}.
\scout's weights are defined on anchor coordinates, so the weight-transfer variants train an aggregator on anchor coordinates, initialized with \scout's weights and a new classifier; either all weights are fine-tuned or only the classifier is trained (linear probe).
\emph{Weight regularization (L2-SP)} adds to fine-tuning an L2 penalty toward \scout's weights ($\lambda{=}0.1$, decayed with $\beta(e)/\beta_0$).
\emph{Slide-embedding distillation} replaces $\mathcal{L}_{\mathrm{att}}$ with a cosine loss between a linear projection of the aggregator's slide embedding and \scout's slide embedding.
The other targets are uniform attention, \scout's attention shuffled across patches within each slide, and the attention of a scratch ABMIL trained with another seed.
The schedules keep $\beta_0$ constant, apply $\beta_0$ in the first epoch only, or decay it linearly over all epochs.
The native-feature counterpart of the weight-transfer comparison is main \cref{fig:entangle}B.

\paragraph{Prior pretraining methods
(main \cref{tab:ablation}).}
Each variant removes or replaces a single component of main \cref{sec:method} and retrains the affected models (the student, and the teachers where the variant changes them) with all other settings unchanged.
The resulting frozen student is then scored zero-shot by lesion AUROC and lesion R-precision on the test slides of \Dann, separately for the 10 seen and the 5 unseen PEs.
Significance is tested with a paired Wilcoxon test over (dataset, PE) pairs, Holm-corrected within each block and column.

\paragraph{Q1, pretraining data.}
Teachers and student are pretrained on TCGA alone or on HistAI alone, with the anchor bank unchanged.

\paragraph{Q2, consensus target.}
\emph{Number of teachers}: the consensus averages the $|\mathcal{T}|\in\{1,2,4\}$ strongest of the eight teachers, ranked by attention precision on the annotated TCGA validation slides (\cref{app:impl}): Virchow2 for $|\mathcal{T}|{=}1$; \{Virchow2, CONCH v1.5\} for $|\mathcal{T}|{=}2$; and \{Virchow2, CONCH v1.5, UNI-2, MUSK\} for $|\mathcal{T}|{=}4$.
The full method uses all eight teachers (\cref{app:teacher_selection}).
\emph{Seeds of one PE}: the consensus averages eight Virchow2 teachers trained with different seeds, matching the number of teachers in the consensus.
\emph{Teacher objective}: teachers are retrained without $\mathcal{L}_{\mathrm{ACMIL}}$, without $\mathcal{L}_{\mathrm{AttriMIL}}$, or with $\mathcal{L}_{\mathrm{CE}}$ only (\cref{app:teacher}), and the student is retrained on their consensus.

\paragraph{Q3, student objective.}
\emph{w/o $\mathcal{L}_{\mathrm{att}}$}: no teacher; the student learns from labels only.
\emph{w/o $\mathcal{L}_{\mathrm{CE}}$}: the student learns from distillation only.
\emph{w/o $\mathcal{L}_{\mathrm{inv}}$}: the three weights of \cref{eq:supp_inv} are set to 0; the two views of a slide are still drawn.
\emph{Teacher losses on the student}: no teacher; the ACMIL terms (five branches) and the AttriMIL terms ($\lambda_{\mathrm{evi}}=2$, $\lambda_{\mathrm{sp}}=0.1$) are applied to the student itself, so that it receives the same objective as a teacher but on anchor coordinates.

\paragraph{Q4, anchor bank.}
\emph{Noise anchors}: the 768 anchor images are replaced by images of uniform RGB noise.
\emph{Pixel-shuffled anchors}: the pixels of each anchor image are randomly permuted, which keeps its color statistics and destroys its tissue structure.
\emph{Natural-image anchors}: the 768 anchor images are replaced by the natural images of \cref{app:cosine}.
\emph{Learned adapter per PE}: anchor coordinates are replaced by a linear map from the L2-normalized features of each seen PE to 768 dimensions, trained jointly with the student; no such map exists for unseen PEs.

\subsection{All benchmarks}
\label{app:perpe}

\Cref{tab:bracs3,tab:perpe_camelyon16,tab:perpe_ubc,tab:perpe_panda,tab:perpe_bcnb_er,tab:perpe_bcnb_pr,tab:perpe_bcnb_her2,tab:cptac_nsclc,tab:cptac_ccrcc,tab:perpe_ebrains_coarse,tab:perpe_ebrains_fine} list the eleven tasks not shown in the main text; BRACS 7-class is main \cref{tab:bracs}.
Every table follows the layout of main \cref{tab:bracs}: the metric of the task (balanced accuracy for BRACS, UBC-OCEAN, and EBRAINS; quadratic weighted kappa for PANDA; accuracy otherwise), averaged over 5 seeds with standard deviations in parentheses; methods grouped into baseline MIL, slide foundation models, and pretrained ABMIL; light gray for the native PE(s) of each method (for ours, the 10 seen PEs); entries of the pretrained-ABMIL block colored \textcolor{gainpos}{green} or \textcolor{gainneg}{red} relative to scratch ABMIL; and the best entry per PE in bold.

\begin{table*}[!t]
\centering
\caption{
\textbf{CAMELYON16 metastasis detection across 15 PEs.}
Accuracy (\%), s.d.\ in parentheses.
Gray: native PE(s) of each method (ours: the 10 seen PEs).
Green/red: above/below scratch ABMIL.
Bold: best per PE.
}
\label{tab:perpe_camelyon16}

\scriptsize
\setlength{\tabcolsep}{2.4pt}
\renewcommand{\arraystretch}{1.07}

\resizebox{\textwidth}{!}{%
\begin{tabular}{
    l
    *{10}{c}
    @{\hspace{3pt}}
    *{5}{c}
}

\toprule

& \multicolumn{10}{c}{Seen during our pretraining}
& \multicolumn{5}{c}{Unseen} \\

\cmidrule(lr){2-11}
\cmidrule(lr){12-16}

Method
& \thead{RN-50\\\cite{he2016deep}}
& \thead{DINO\\\cite{caron2021emerging}}
& \thead{CONCH\\\cite{lu2024visual}}
& \thead{UNI\\\cite{chen2024uni}}
& \thead{Hibou-B\\\cite{nechaev2024hibou}}
& \thead{Virch2\\\cite{zimmermann2024virchow2}}
& \thead{Phikon2\\\cite{filiot2024phikonv2}}
& \thead{MUSK\\\cite{xiang2025musk}}
& \thead{CON1.5\\\cite{ding2025titan}}
& \thead{UNI2\\\cite{chen2024uni}}
& \thead{CTrans\\\cite{wang2022ctranspath}}
& \thead{Kaiko-S\\\cite{ai2024largescale}}
& \thead{GigaP\\\cite{xu2024gigapath}}
& \thead{H-Opt1\\\cite{scalbert2026hoptimus1}}
& \thead{GenBio\\\cite{kapse2026genbiopathfm}} \\

\midrule

\multicolumn{16}{l}{\textit{Baseline MIL}} \\

DTFD-MIL~\cite{zhang2022dtfd}
& \mstd{\textbf{77.1}}{2.0}
& \mstd{89.6}{2.3}
& \mstd{\textbf{94.6}}{1.1}
& \mstd{95.2}{3.3}
& \mstd{95.8}{1.3}
& \mstd{96.0}{2.3}
& \mstd{88.5}{1.1}
& \mstd{91.3}{1.8}
& \mstd{93.5}{1.3}
& \mstd{96.6}{0.9}
& \mstd{85.7}{3.1}
& \mstd{87.6}{3.3}
& \mstd{95.3}{0.8}
& \mstd{\textbf{96.4}}{0.7}
& \mstd{95.3}{0.8}
\\

TransMIL~\cite{shao2021transmil}
& \mstd{67.0}{8.0}
& \mstd{87.0}{2.4}
& \mstd{92.4}{2.3}
& \mstd{90.7}{3.1}
& \mstd{88.7}{8.4}
& \mstd{96.3}{1.9}
& \mstd{82.3}{4.9}
& \mstd{90.1}{2.8}
& \mstd{92.7}{1.2}
& \mstd{95.3}{1.3}
& \mstd{78.3}{2.4}
& \mstd{86.0}{3.1}
& \mstd{94.7}{1.3}
& \mstd{92.2}{3.9}
& \mstd{95.3}{0.9}
\\

WiKG~\cite{li2024dynamic}
& \mstd{69.5}{6.7}
& \mstd{64.0}{11.0}
& \mstd{90.7}{3.0}
& \mstd{80.2}{16.3}
& \mstd{84.3}{11.2}
& \mstd{96.7}{1.3}
& \mstd{87.1}{2.2}
& \mstd{75.5}{15.7}
& \mstd{90.1}{10.4}
& \mstd{95.7}{0.7}
& \mstd{82.8}{3.2}
& \mstd{56.0}{10.1}
& \mstd{90.5}{10.1}
& \mstd{85.9}{12.5}
& \mstd{92.2}{7.2}
\\
\bottomrule
\end{tabular}%
}

\end{table*}

\begin{table*}[!t]
\centering
\caption{
\textbf{UBC-OCEAN 5-class subtyping across 15 PEs.}
Balanced accuracy (\%), s.d.\ in parentheses.
Gray: native PE(s) of each method (ours: the 10 seen PEs).
Green/red: above/below scratch ABMIL.
Bold: best per PE.
}
\label{tab:perpe_ubc}

\scriptsize
\setlength{\tabcolsep}{2.4pt}
\renewcommand{\arraystretch}{1.07}

\resizebox{\textwidth}{!}{%
\begin{tabular}{
    l
    *{10}{c}
    @{\hspace{3pt}}
    *{5}{c}
}

\toprule

& \multicolumn{10}{c}{Seen during our pretraining}
& \multicolumn{5}{c}{Unseen} \\

\cmidrule(lr){2-11}
\cmidrule(lr){12-16}

Method
& \thead{RN-50\\\cite{he2016deep}}
& \thead{DINO\\\cite{caron2021emerging}}
& \thead{CONCH\\\cite{lu2024visual}}
& \thead{UNI\\\cite{chen2024uni}}
& \thead{Hibou-B\\\cite{nechaev2024hibou}}
& \thead{Virch2\\\cite{zimmermann2024virchow2}}
& \thead{Phikon2\\\cite{filiot2024phikonv2}}
& \thead{MUSK\\\cite{xiang2025musk}}
& \thead{CON1.5\\\cite{ding2025titan}}
& \thead{UNI2\\\cite{chen2024uni}}
& \thead{CTrans\\\cite{wang2022ctranspath}}
& \thead{Kaiko-S\\\cite{ai2024largescale}}
& \thead{GigaP\\\cite{xu2024gigapath}}
& \thead{H-Opt1\\\cite{scalbert2026hoptimus1}}
& \thead{GenBio\\\cite{kapse2026genbiopathfm}} \\

\midrule

\multicolumn{16}{l}{\textit{Baseline MIL}} \\

DTFD-MIL~\cite{zhang2022dtfd}
& \mstd{\textbf{79.2}}{3.4}
& \mstd{82.2}{3.2}
& \mstd{87.3}{2.5}
& \mstd{86.5}{2.0}
& \mstd{83.5}{2.4}
& \mstd{86.8}{2.0}
& \mstd{82.8}{1.7}
& \mstd{84.6}{2.7}
& \mstd{86.5}{2.0}
& \mstd{88.6}{2.5}
& \mstd{81.5}{2.5}
& \mstd{81.0}{4.0}
& \mstd{85.8}{3.5}
& \mstd{84.6}{2.4}
& \mstd{84.5}{2.0}
\\

TransMIL~\cite{shao2021transmil}
& \mstd{63.8}{6.6}
& \mstd{83.0}{2.9}
& \mstd{81.9}{3.2}
& \mstd{85.0}{1.5}
& \mstd{82.5}{3.7}
& \mstd{88.1}{2.2}
& \mstd{81.0}{3.8}
& \mstd{84.5}{3.3}
& \mstd{87.6}{2.0}
& \mstd{86.9}{2.3}
& \mstd{80.0}{3.9}
& \mstd{83.0}{2.3}
& \mstd{84.9}{2.4}
& \mstd{84.9}{1.9}
& \mstd{86.1}{2.5}
\\

WiKG~\cite{li2024dynamic}
& \mstd{71.9}{3.6}
& \mstd{81.0}{3.0}
& \mstd{85.3}{2.0}
& \mstd{85.7}{2.4}
& \mstd{82.8}{2.5}
& \mstd{87.1}{2.6}
& \mstd{83.3}{3.0}
& \mstd{83.3}{2.3}
& \mstd{86.3}{3.4}
& \mstd{87.7}{1.9}
& \mstd{76.8}{2.9}
& \mstd{78.0}{4.3}
& \mstd{86.0}{3.3}
& \mstd{85.0}{3.6}
& \mstd{85.8}{2.1}
\\
\bottomrule
\end{tabular}%
}

\end{table*}

\begin{table*}[!t]
\centering
\caption{
\textbf{PANDA ISUP grading across 15 PEs.}
Quadratic weighted kappa ($\times$100), s.d.\ in parentheses.
Gray: native PE(s) of each method (ours: the 10 seen PEs).
Green/red: above/below scratch ABMIL.
Bold: best per PE.
}
\label{tab:perpe_panda}

\scriptsize
\setlength{\tabcolsep}{2.4pt}
\renewcommand{\arraystretch}{1.07}

\resizebox{\textwidth}{!}{%
\begin{tabular}{
    l
    *{10}{c}
    @{\hspace{3pt}}
    *{5}{c}
}

\toprule

& \multicolumn{10}{c}{Seen during our pretraining}
& \multicolumn{5}{c}{Unseen} \\

\cmidrule(lr){2-11}
\cmidrule(lr){12-16}

Method
& \thead{RN-50\\\cite{he2016deep}}
& \thead{DINO\\\cite{caron2021emerging}}
& \thead{CONCH\\\cite{lu2024visual}}
& \thead{UNI\\\cite{chen2024uni}}
& \thead{Hibou-B\\\cite{nechaev2024hibou}}
& \thead{Virch2\\\cite{zimmermann2024virchow2}}
& \thead{Phikon2\\\cite{filiot2024phikonv2}}
& \thead{MUSK\\\cite{xiang2025musk}}
& \thead{CON1.5\\\cite{ding2025titan}}
& \thead{UNI2\\\cite{chen2024uni}}
& \thead{CTrans\\\cite{wang2022ctranspath}}
& \thead{Kaiko-S\\\cite{ai2024largescale}}
& \thead{GigaP\\\cite{xu2024gigapath}}
& \thead{H-Opt1\\\cite{scalbert2026hoptimus1}}
& \thead{GenBio\\\cite{kapse2026genbiopathfm}} \\

\midrule

\multicolumn{16}{l}{\textit{Baseline MIL}} \\

DTFD-MIL~\cite{zhang2022dtfd}
& \mstd{\textbf{77.4}}{0.7}
& \mstd{82.2}{0.7}
& \mstd{84.3}{0.6}
& \mstd{\textbf{86.8}}{0.5}
& \mstd{84.2}{0.3}
& \mstd{85.5}{0.4}
& \mstd{\textbf{83.8}}{0.8}
& \mstd{81.4}{0.8}
& \mstd{82.9}{0.6}
& \mstd{85.7}{0.5}
& \mstd{82.1}{1.5}
& \mstd{\textbf{84.5}}{0.6}
& \mstd{83.9}{0.2}
& \mstd{\textbf{86.1}}{1.2}
& \mstd{\textbf{86.1}}{1.0}
\\

TransMIL~\cite{shao2021transmil}
& \mstd{71.8}{2.0}
& \mstd{79.0}{1.2}
& \mstd{81.2}{0.8}
& \mstd{83.1}{1.2}
& \mstd{80.3}{1.1}
& \mstd{83.5}{0.9}
& \mstd{78.6}{1.3}
& \mstd{78.6}{1.0}
& \mstd{81.8}{0.8}
& \mstd{83.8}{0.8}
& \mstd{79.1}{0.9}
& \mstd{81.1}{1.7}
& \mstd{82.0}{0.4}
& \mstd{83.4}{0.6}
& \mstd{83.7}{0.4}
\\

WiKG~\cite{li2024dynamic}
& \mstd{76.9}{1.5}
& \mstd{79.9}{0.4}
& \mstd{82.6}{0.7}
& \mstd{83.2}{2.0}
& \mstd{82.0}{0.7}
& \mstd{83.8}{0.3}
& \mstd{82.3}{0.8}
& \mstd{79.2}{0.4}
& \mstd{82.1}{0.6}
& \mstd{84.6}{0.4}
& \mstd{82.1}{0.6}
& \mstd{80.3}{0.8}
& \mstd{83.0}{0.7}
& \mstd{83.4}{1.2}
& \mstd{85.0}{0.3}
\\
\bottomrule
\end{tabular}%
}

\end{table*}

\begin{table*}[!t]
\centering
\caption{
\textbf{BCNB ER status prediction across 15 PEs.}
Accuracy (\%), s.d.\ in parentheses.
Gray: native PE(s) of each method (ours: the 10 seen PEs).
Green/red: above/below scratch ABMIL.
Bold: best per PE.
}
\label{tab:perpe_bcnb_er}

\scriptsize
\setlength{\tabcolsep}{2.4pt}
\renewcommand{\arraystretch}{1.07}

\resizebox{\textwidth}{!}{%
\begin{tabular}{
    l
    *{10}{c}
    @{\hspace{3pt}}
    *{5}{c}
}

\toprule

& \multicolumn{10}{c}{Seen during our pretraining}
& \multicolumn{5}{c}{Unseen} \\

\cmidrule(lr){2-11}
\cmidrule(lr){12-16}

Method
& \thead{RN-50\\\cite{he2016deep}}
& \thead{DINO\\\cite{caron2021emerging}}
& \thead{CONCH\\\cite{lu2024visual}}
& \thead{UNI\\\cite{chen2024uni}}
& \thead{Hibou-B\\\cite{nechaev2024hibou}}
& \thead{Virch2\\\cite{zimmermann2024virchow2}}
& \thead{Phikon2\\\cite{filiot2024phikonv2}}
& \thead{MUSK\\\cite{xiang2025musk}}
& \thead{CON1.5\\\cite{ding2025titan}}
& \thead{UNI2\\\cite{chen2024uni}}
& \thead{CTrans\\\cite{wang2022ctranspath}}
& \thead{Kaiko-S\\\cite{ai2024largescale}}
& \thead{GigaP\\\cite{xu2024gigapath}}
& \thead{H-Opt1\\\cite{scalbert2026hoptimus1}}
& \thead{GenBio\\\cite{kapse2026genbiopathfm}} \\

\midrule

\multicolumn{16}{l}{\textit{Baseline MIL}} \\

DTFD-MIL~\cite{zhang2022dtfd}
& \mstd{70.0}{3.9}
& \mstd{\textbf{83.1}}{2.1}
& \mstd{84.5}{2.1}
& \mstd{84.1}{2.2}
& \mstd{82.4}{2.5}
& \mstd{82.7}{2.0}
& \mstd{79.9}{6.8}
& \mstd{80.4}{4.7}
& \mstd{82.0}{2.8}
& \mstd{83.6}{2.1}
& \mstd{79.7}{3.9}
& \mstd{82.0}{2.7}
& \mstd{82.6}{4.7}
& \mstd{84.4}{2.5}
& \mstd{83.2}{2.1}
\\

TransMIL~\cite{shao2021transmil}
& \mstd{70.6}{7.6}
& \mstd{79.2}{3.7}
& \mstd{81.0}{4.2}
& \mstd{83.2}{2.4}
& \mstd{78.5}{3.4}
& \mstd{81.8}{1.7}
& \mstd{79.3}{2.2}
& \mstd{79.4}{2.1}
& \mstd{79.4}{4.4}
& \mstd{84.0}{1.8}
& \mstd{77.0}{3.0}
& \mstd{80.7}{2.3}
& \mstd{82.5}{2.0}
& \mstd{84.1}{1.7}
& \mstd{80.0}{6.2}
\\

WiKG~\cite{li2024dynamic}
& \mstd{75.5}{3.3}
& \mstd{76.8}{3.5}
& \mstd{80.9}{2.7}
& \mstd{83.8}{0.8}
& \mstd{76.6}{7.1}
& \mstd{82.4}{2.1}
& \mstd{81.9}{2.0}
& \mstd{71.8}{9.6}
& \mstd{77.2}{3.3}
& \mstd{83.8}{3.9}
& \mstd{78.1}{6.6}
& \mstd{74.5}{8.4}
& \mstd{82.7}{2.5}
& \mstd{82.6}{4.7}
& \mstd{82.9}{2.5}
\\
\bottomrule
\end{tabular}%
}

\end{table*}

\begin{table*}[!t]
\centering
\caption{
\textbf{BCNB PR status prediction across 15 PEs.}
Accuracy (\%), s.d.\ in parentheses.
Gray: native PE(s) of each method (ours: the 10 seen PEs).
Green/red: above/below scratch ABMIL.
Bold: best per PE.
}
\label{tab:perpe_bcnb_pr}

\scriptsize
\setlength{\tabcolsep}{2.4pt}
\renewcommand{\arraystretch}{1.07}

\resizebox{\textwidth}{!}{%
\begin{tabular}{
    l
    *{10}{c}
    @{\hspace{3pt}}
    *{5}{c}
}

\toprule

& \multicolumn{10}{c}{Seen during our pretraining}
& \multicolumn{5}{c}{Unseen} \\

\cmidrule(lr){2-11}
\cmidrule(lr){12-16}

Method
& \thead{RN-50\\\cite{he2016deep}}
& \thead{DINO\\\cite{caron2021emerging}}
& \thead{CONCH\\\cite{lu2024visual}}
& \thead{UNI\\\cite{chen2024uni}}
& \thead{Hibou-B\\\cite{nechaev2024hibou}}
& \thead{Virch2\\\cite{zimmermann2024virchow2}}
& \thead{Phikon2\\\cite{filiot2024phikonv2}}
& \thead{MUSK\\\cite{xiang2025musk}}
& \thead{CON1.5\\\cite{ding2025titan}}
& \thead{UNI2\\\cite{chen2024uni}}
& \thead{CTrans\\\cite{wang2022ctranspath}}
& \thead{Kaiko-S\\\cite{ai2024largescale}}
& \thead{GigaP\\\cite{xu2024gigapath}}
& \thead{H-Opt1\\\cite{scalbert2026hoptimus1}}
& \thead{GenBio\\\cite{kapse2026genbiopathfm}} \\

\midrule

\multicolumn{16}{l}{\textit{Baseline MIL}} \\

DTFD-MIL~\cite{zhang2022dtfd}
& \mstd{68.4}{9.0}
& \mstd{78.8}{3.2}
& \mstd{75.7}{7.8}
& \mstd{75.0}{9.5}
& \mstd{78.1}{3.1}
& \mstd{75.3}{9.8}
& \mstd{75.0}{6.3}
& \mstd{76.1}{4.3}
& \mstd{77.2}{6.6}
& \mstd{\textbf{80.4}}{3.7}
& \mstd{75.8}{3.3}
& \mstd{\textbf{80.5}}{2.3}
& \mstd{79.0}{2.0}
& \mstd{79.9}{4.1}
& \mstd{78.4}{3.2}
\\

TransMIL~\cite{shao2021transmil}
& \mstd{64.5}{8.5}
& \mstd{76.6}{3.7}
& \mstd{75.1}{2.7}
& \mstd{79.3}{2.5}
& \mstd{77.7}{2.7}
& \mstd{78.9}{1.9}
& \mstd{75.0}{4.1}
& \mstd{75.0}{4.1}
& \mstd{75.9}{3.1}
& \mstd{77.4}{6.8}
& \mstd{74.5}{2.2}
& \mstd{76.2}{3.3}
& \mstd{74.9}{5.5}
& \mstd{78.8}{1.1}
& \mstd{74.6}{8.0}
\\

WiKG~\cite{li2024dynamic}
& \mstd{70.8}{6.4}
& \mstd{69.1}{6.7}
& \mstd{74.7}{7.0}
& \mstd{75.9}{6.5}
& \mstd{70.1}{7.6}
& \mstd{79.6}{3.0}
& \mstd{74.0}{6.3}
& \mstd{71.4}{9.4}
& \mstd{74.1}{6.5}
& \mstd{78.2}{4.5}
& \mstd{68.4}{10.9}
& \mstd{71.6}{8.7}
& \mstd{77.1}{3.4}
& \mstd{79.7}{2.9}
& \mstd{78.3}{1.6}
\\
\bottomrule
\end{tabular}%
}

\end{table*}

\begin{table*}[!t]
\centering
\caption{
\textbf{BCNB HER2 status prediction across 15 PEs.}
Accuracy (\%), s.d.\ in parentheses.
Gray: native PE(s) of each method (ours: the 10 seen PEs).
Green/red: above/below scratch ABMIL.
Bold: best per PE.
}
\label{tab:perpe_bcnb_her2}

\scriptsize
\setlength{\tabcolsep}{2.4pt}
\renewcommand{\arraystretch}{1.07}

\resizebox{\textwidth}{!}{%
\begin{tabular}{
    l
    *{10}{c}
    @{\hspace{3pt}}
    *{5}{c}
}

\toprule

& \multicolumn{10}{c}{Seen during our pretraining}
& \multicolumn{5}{c}{Unseen} \\

\cmidrule(lr){2-11}
\cmidrule(lr){12-16}

Method
& \thead{RN-50\\\cite{he2016deep}}
& \thead{DINO\\\cite{caron2021emerging}}
& \thead{CONCH\\\cite{lu2024visual}}
& \thead{UNI\\\cite{chen2024uni}}
& \thead{Hibou-B\\\cite{nechaev2024hibou}}
& \thead{Virch2\\\cite{zimmermann2024virchow2}}
& \thead{Phikon2\\\cite{filiot2024phikonv2}}
& \thead{MUSK\\\cite{xiang2025musk}}
& \thead{CON1.5\\\cite{ding2025titan}}
& \thead{UNI2\\\cite{chen2024uni}}
& \thead{CTrans\\\cite{wang2022ctranspath}}
& \thead{Kaiko-S\\\cite{ai2024largescale}}
& \thead{GigaP\\\cite{xu2024gigapath}}
& \thead{H-Opt1\\\cite{scalbert2026hoptimus1}}
& \thead{GenBio\\\cite{kapse2026genbiopathfm}} \\

\midrule

\multicolumn{16}{l}{\textit{Baseline MIL}} \\

DTFD-MIL~\cite{zhang2022dtfd}
& \mstd{63.4}{10.2}
& \mstd{65.9}{13.2}
& \mstd{68.5}{9.6}
& \mstd{71.0}{8.6}
& \mstd{66.7}{11.9}
& \mstd{69.4}{8.9}
& \mstd{69.0}{8.1}
& \mstd{67.8}{9.5}
& \mstd{69.8}{8.1}
& \mstd{73.3}{9.5}
& \mstd{59.0}{8.7}
& \mstd{63.8}{11.7}
& \mstd{68.4}{9.9}
& \mstd{71.1}{10.6}
& \mstd{63.7}{13.0}
\\

TransMIL~\cite{shao2021transmil}
& \mstd{52.8}{18.4}
& \mstd{70.9}{4.6}
& \mstd{69.3}{9.7}
& \mstd{72.0}{2.7}
& \mstd{65.6}{6.1}
& \mstd{71.4}{2.1}
& \mstd{65.0}{9.5}
& \mstd{68.3}{9.1}
& \mstd{62.2}{10.8}
& \mstd{71.7}{3.9}
& \mstd{63.5}{13.8}
& \mstd{71.3}{4.0}
& \mstd{73.8}{2.1}
& \mstd{71.7}{5.1}
& \mstd{71.4}{4.8}
\\

WiKG~\cite{li2024dynamic}
& \mstd{52.8}{15.9}
& \mstd{64.4}{11.7}
& \mstd{67.8}{7.2}
& \mstd{60.0}{10.6}
& \mstd{62.4}{13.0}
& \mstd{63.2}{11.5}
& \mstd{62.2}{14.5}
& \mstd{62.6}{14.4}
& \mstd{61.8}{15.7}
& \mstd{\textbf{75.6}}{4.6}
& \mstd{64.4}{9.6}
& \mstd{57.5}{14.0}
& \mstd{58.9}{15.8}
& \mstd{63.8}{14.8}
& \mstd{66.9}{9.3}
\\
\bottomrule
\end{tabular}%
}

\end{table*}

\begin{table*}[!t]
\centering
\caption{
\textbf{CPTAC NSCLC subtyping across 15 PEs.}
Accuracy (\%), s.d.\ in parentheses.
Gray: native PE(s) of each method (ours: the 10 seen PEs).
Green/red: above/below scratch ABMIL.
Bold: best per PE.
}
\label{tab:cptac_nsclc}

\scriptsize
\setlength{\tabcolsep}{2.4pt}
\renewcommand{\arraystretch}{1.07}

\resizebox{\textwidth}{!}{%
\begin{tabular}{
    l
    *{10}{c}
    @{\hspace{3pt}}
    *{5}{c}
}

\toprule

& \multicolumn{10}{c}{Seen during our pretraining}
& \multicolumn{5}{c}{Unseen} \\

\cmidrule(lr){2-11}
\cmidrule(lr){12-16}

Method
& \thead{RN-50\\\cite{he2016deep}}
& \thead{DINO\\\cite{caron2021emerging}}
& \thead{CONCH\\\cite{lu2024visual}}
& \thead{UNI\\\cite{chen2024uni}}
& \thead{Hibou-B\\\cite{nechaev2024hibou}}
& \thead{Virch2\\\cite{zimmermann2024virchow2}}
& \thead{Phikon2\\\cite{filiot2024phikonv2}}
& \thead{MUSK\\\cite{xiang2025musk}}
& \thead{CON1.5\\\cite{ding2025titan}}
& \thead{UNI2\\\cite{chen2024uni}}
& \thead{CTrans\\\cite{wang2022ctranspath}}
& \thead{Kaiko-S\\\cite{ai2024largescale}}
& \thead{GigaP\\\cite{xu2024gigapath}}
& \thead{H-Opt1\\\cite{scalbert2026hoptimus1}}
& \thead{GenBio\\\cite{kapse2026genbiopathfm}} \\

\midrule

\multicolumn{16}{l}{\textit{Baseline MIL}} \\

DTFD-MIL~\cite{zhang2022dtfd}
& \mstd{61.3}{19.6}
& \mstd{75.4}{7.0}
& \mstd{86.4}{4.8}
& \mstd{82.5}{5.5}
& \mstd{74.0}{14.6}
& \mstd{86.1}{6.3}
& \mstd{73.0}{12.7}
& \mstd{73.6}{8.5}
& \mstd{85.5}{8.0}
& \mstd{82.8}{8.5}
& \mstd{70.7}{6.7}
& \mstd{74.8}{10.2}
& \mstd{76.3}{14.0}
& \mstd{81.9}{6.9}
& \mstd{\textbf{91.7}}{4.1}
\\

TransMIL~\cite{shao2021transmil}
& \mstd{69.3}{8.7}
& \mstd{79.9}{7.3}
& \mstd{85.1}{7.2}
& \mstd{77.9}{8.0}
& \mstd{71.7}{13.6}
& \mstd{82.1}{10.6}
& \mstd{67.2}{11.7}
& \mstd{77.7}{7.8}
& \mstd{86.1}{8.3}
& \mstd{76.2}{14.5}
& \mstd{72.9}{9.7}
& \mstd{72.9}{9.8}
& \mstd{74.7}{12.8}
& \mstd{82.7}{8.1}
& \mstd{85.9}{8.7}
\\

WiKG~\cite{li2024dynamic}
& \mstd{66.3}{13.5}
& \mstd{79.0}{6.7}
& \mstd{87.3}{6.7}
& \mstd{78.7}{9.6}
& \mstd{59.9}{9.1}
& \mstd{79.0}{8.5}
& \mstd{76.0}{9.8}
& \mstd{72.4}{5.5}
& \mstd{85.8}{8.7}
& \mstd{82.2}{10.8}
& \mstd{69.7}{6.9}
& \mstd{68.0}{13.4}
& \mstd{74.5}{11.9}
& \mstd{78.4}{11.9}
& \mstd{83.4}{8.7}
\\
\bottomrule
\end{tabular}%
}

\end{table*}

\begin{table*}[!t]
\centering
\caption{
\textbf{CPTAC CCRCC grading across 15 PEs.}
Accuracy (\%), s.d.\ in parentheses.
Gray: native PE(s) of each method (ours: the 10 seen PEs).
Green/red: above/below scratch ABMIL.
Bold: best per PE.
}
\label{tab:cptac_ccrcc}

\scriptsize
\setlength{\tabcolsep}{2.4pt}
\renewcommand{\arraystretch}{1.07}

\resizebox{\textwidth}{!}{%
\begin{tabular}{
    l
    *{10}{c}
    @{\hspace{3pt}}
    *{5}{c}
}

\toprule

& \multicolumn{10}{c}{Seen during our pretraining}
& \multicolumn{5}{c}{Unseen} \\

\cmidrule(lr){2-11}
\cmidrule(lr){12-16}

Method
& \thead{RN-50\\\cite{he2016deep}}
& \thead{DINO\\\cite{caron2021emerging}}
& \thead{CONCH\\\cite{lu2024visual}}
& \thead{UNI\\\cite{chen2024uni}}
& \thead{Hibou-B\\\cite{nechaev2024hibou}}
& \thead{Virch2\\\cite{zimmermann2024virchow2}}
& \thead{Phikon2\\\cite{filiot2024phikonv2}}
& \thead{MUSK\\\cite{xiang2025musk}}
& \thead{CON1.5\\\cite{ding2025titan}}
& \thead{UNI2\\\cite{chen2024uni}}
& \thead{CTrans\\\cite{wang2022ctranspath}}
& \thead{Kaiko-S\\\cite{ai2024largescale}}
& \thead{GigaP\\\cite{xu2024gigapath}}
& \thead{H-Opt1\\\cite{scalbert2026hoptimus1}}
& \thead{GenBio\\\cite{kapse2026genbiopathfm}} \\

\midrule

\multicolumn{16}{l}{\textit{Baseline MIL}} \\

DTFD-MIL~\cite{zhang2022dtfd}
& \mstd{58.5}{8.8}
& \mstd{67.6}{9.7}
& \mstd{\textbf{72.1}}{3.4}
& \mstd{67.6}{5.0}
& \mstd{64.4}{11.9}
& \mstd{67.2}{4.0}
& \mstd{64.4}{4.9}
& \mstd{64.4}{8.3}
& \mstd{68.9}{5.8}
& \mstd{68.5}{5.5}
& \mstd{65.3}{5.5}
& \mstd{66.7}{2.1}
& \mstd{67.5}{2.9}
& \mstd{65.3}{4.6}
& \mstd{64.9}{6.9}
\\

TransMIL~\cite{shao2021transmil}
& \mstd{57.2}{9.9}
& \mstd{65.8}{9.8}
& \mstd{68.9}{8.1}
& \mstd{68.5}{6.7}
& \mstd{61.0}{10.3}
& \mstd{68.5}{5.8}
& \mstd{66.4}{7.5}
& \mstd{65.8}{7.5}
& \mstd{67.8}{6.7}
& \mstd{67.3}{4.9}
& \mstd{63.3}{8.4}
& \mstd{70.0}{7.2}
& \mstd{66.9}{4.8}
& \mstd{69.6}{7.7}
& \mstd{66.4}{10.1}
\\

WiKG~\cite{li2024dynamic}
& \mstd{65.8}{5.0}
& \mstd{64.4}{4.9}
& \mstd{69.8}{6.9}
& \mstd{69.4}{2.6}
& \mstd{67.7}{11.7}
& \mstd{64.8}{10.6}
& \mstd{65.8}{2.9}
& \mstd{64.9}{4.1}
& \mstd{70.3}{5.9}
& \mstd{68.0}{3.2}
& \mstd{62.6}{4.4}
& \mstd{67.2}{8.7}
& \mstd{64.4}{5.7}
& \mstd{72.6}{6.8}
& \mstd{67.6}{5.4}
\\
\bottomrule
\end{tabular}%
}

\end{table*}

\begin{table*}[!t]
\centering
\caption{
\textbf{EBRAINS 12-class subtyping across 15 PEs.}
Balanced accuracy (\%), s.d.\ in parentheses.
Gray: native PE(s) of each method (ours: the 10 seen PEs).
Green/red: above/below scratch ABMIL.
Bold: best per PE.
}
\label{tab:perpe_ebrains_coarse}

\scriptsize
\setlength{\tabcolsep}{2.4pt}
\renewcommand{\arraystretch}{1.07}

\resizebox{\textwidth}{!}{%
\begin{tabular}{
    l
    *{10}{c}
    @{\hspace{3pt}}
    *{5}{c}
}

\toprule

& \multicolumn{10}{c}{Seen during our pretraining}
& \multicolumn{5}{c}{Unseen} \\

\cmidrule(lr){2-11}
\cmidrule(lr){12-16}

Method
& \thead{RN-50\\\cite{he2016deep}}
& \thead{DINO\\\cite{caron2021emerging}}
& \thead{CONCH\\\cite{lu2024visual}}
& \thead{UNI\\\cite{chen2024uni}}
& \thead{Hibou-B\\\cite{nechaev2024hibou}}
& \thead{Virch2\\\cite{zimmermann2024virchow2}}
& \thead{Phikon2\\\cite{filiot2024phikonv2}}
& \thead{MUSK\\\cite{xiang2025musk}}
& \thead{CON1.5\\\cite{ding2025titan}}
& \thead{UNI2\\\cite{chen2024uni}}
& \thead{CTrans\\\cite{wang2022ctranspath}}
& \thead{Kaiko-S\\\cite{ai2024largescale}}
& \thead{GigaP\\\cite{xu2024gigapath}}
& \thead{H-Opt1\\\cite{scalbert2026hoptimus1}}
& \thead{GenBio\\\cite{kapse2026genbiopathfm}} \\

\midrule

\multicolumn{16}{l}{\textit{Baseline MIL}} \\

DTFD-MIL~\cite{zhang2022dtfd}
& \mstd{68.6}{1.9}
& \mstd{86.8}{0.9}
& \mstd{89.9}{0.5}
& \mstd{89.2}{0.1}
& \mstd{81.6}{2.2}
& \mstd{89.7}{1.1}
& \mstd{78.8}{0.1}
& \mstd{86.4}{3.0}
& \mstd{\textbf{92.2}}{0.9}
& \mstd{88.1}{1.1}
& \mstd{82.0}{4.2}
& \mstd{82.9}{1.8}
& \mstd{88.4}{0.7}
& \mstd{87.7}{0.4}
& \mstd{86.9}{1.7}
\\

TransMIL~\cite{shao2021transmil}
& \mstd{62.8}{3.3}
& \mstd{80.8}{2.8}
& \mstd{86.4}{0.2}
& \mstd{82.6}{0.1}
& \mstd{82.2}{1.7}
& \mstd{84.8}{1.9}
& \mstd{76.8}{2.5}
& \mstd{78.6}{0.6}
& \mstd{88.9}{0.5}
& \mstd{89.7}{2.4}
& \mstd{76.4}{2.3}
& \mstd{78.0}{5.9}
& \mstd{82.6}{0.3}
& \mstd{83.4}{3.3}
& \mstd{85.4}{0.5}
\\

WiKG~\cite{li2024dynamic}
& \mstd{\textbf{73.6}}{2.7}
& \mstd{79.7}{0.0}
& \mstd{90.2}{0.5}
& \mstd{84.0}{0.0}
& \mstd{81.8}{1.3}
& \mstd{88.4}{1.0}
& \mstd{\textbf{82.7}}{0.1}
& \mstd{84.8}{2.1}
& \mstd{90.3}{2.7}
& \mstd{91.2}{1.2}
& \mstd{79.0}{0.2}
& \mstd{77.4}{0.2}
& \mstd{86.3}{2.7}
& \mstd{\textbf{88.9}}{0.6}
& \mstd{87.1}{0.4}
\\
\bottomrule
\end{tabular}%
}

\end{table*}

\begin{table*}[!t]
\centering
\caption{
\textbf{EBRAINS 30-class subtyping across 15 PEs.}
Balanced accuracy (\%), s.d.\ in parentheses.
Gray: native PE(s) of each method (ours: the 10 seen PEs).
Green/red: above/below scratch ABMIL.
Bold: best per PE.
}
\label{tab:perpe_ebrains_fine}

\scriptsize
\setlength{\tabcolsep}{2.4pt}
\renewcommand{\arraystretch}{1.07}

\resizebox{\textwidth}{!}{%
\begin{tabular}{
    l
    *{10}{c}
    @{\hspace{3pt}}
    *{5}{c}
}

\toprule

& \multicolumn{10}{c}{Seen during our pretraining}
& \multicolumn{5}{c}{Unseen} \\

\cmidrule(lr){2-11}
\cmidrule(lr){12-16}

Method
& \thead{RN-50\\\cite{he2016deep}}
& \thead{DINO\\\cite{caron2021emerging}}
& \thead{CONCH\\\cite{lu2024visual}}
& \thead{UNI\\\cite{chen2024uni}}
& \thead{Hibou-B\\\cite{nechaev2024hibou}}
& \thead{Virch2\\\cite{zimmermann2024virchow2}}
& \thead{Phikon2\\\cite{filiot2024phikonv2}}
& \thead{MUSK\\\cite{xiang2025musk}}
& \thead{CON1.5\\\cite{ding2025titan}}
& \thead{UNI2\\\cite{chen2024uni}}
& \thead{CTrans\\\cite{wang2022ctranspath}}
& \thead{Kaiko-S\\\cite{ai2024largescale}}
& \thead{GigaP\\\cite{xu2024gigapath}}
& \thead{H-Opt1\\\cite{scalbert2026hoptimus1}}
& \thead{GenBio\\\cite{kapse2026genbiopathfm}} \\

\midrule

\multicolumn{16}{l}{\textit{Baseline MIL}} \\

DTFD-MIL~\cite{zhang2022dtfd}
& \mstd{51.7}{2.6}
& \mstd{62.5}{1.3}
& \mstd{67.3}{0.2}
& \mstd{65.8}{2.2}
& \mstd{64.8}{2.6}
& \mstd{71.0}{0.8}
& \mstd{62.6}{2.2}
& \mstd{64.6}{0.8}
& \mstd{69.8}{0.3}
& \mstd{69.8}{2.3}
& \mstd{57.9}{3.4}
& \mstd{64.4}{0.4}
& \mstd{65.5}{1.5}
& \mstd{68.2}{0.0}
& \mstd{70.8}{0.4}
\\

TransMIL~\cite{shao2021transmil}
& \mstd{39.8}{0.7}
& \mstd{59.2}{1.1}
& \mstd{66.4}{2.0}
& \mstd{66.9}{2.0}
& \mstd{58.8}{0.3}
& \mstd{67.9}{0.3}
& \mstd{56.0}{0.3}
& \mstd{62.0}{2.8}
& \mstd{66.8}{0.7}
& \mstd{67.5}{0.2}
& \mstd{56.4}{4.6}
& \mstd{61.2}{2.8}
& \mstd{66.7}{1.4}
& \mstd{66.0}{2.0}
& \mstd{64.4}{0.5}
\\

WiKG~\cite{li2024dynamic}
& \mstd{53.5}{0.2}
& \mstd{63.1}{4.0}
& \mstd{68.2}{0.1}
& \mstd{67.6}{0.3}
& \mstd{62.3}{2.0}
& \mstd{\textbf{72.4}}{2.3}
& \mstd{61.5}{1.5}
& \mstd{62.5}{0.8}
& \mstd{71.2}{0.2}
& \mstd{69.7}{0.6}
& \mstd{60.8}{1.8}
& \mstd{57.1}{0.4}
& \mstd{65.2}{1.4}
& \mstd{68.1}{1.9}
& \mstd{68.7}{0.7}
\\
\bottomrule
\end{tabular}%
}

\end{table*}

\end{document}